\documentclass{article} 
\usepackage[final]{colm2026_conference}

\usepackage{microtype}
\usepackage{hyperref}
\usepackage{url}
\usepackage{booktabs}
\usepackage{caption}
\usepackage{subcaption}
\usepackage{xcolor}
\usepackage{mdframed}
\usepackage[pass]{geometry}
\usepackage{graphicx} 
\usepackage{enumitem}

\definecolor{darkblue}{rgb}{0, 0, 0.5}
\definecolor{lightgray}{RGB}{240,240,240}
\newmdenv[
  backgroundcolor=lightgray,
  linecolor=lightgray,
  innerleftmargin=12pt,
  innerrightmargin=12pt,
  innertopmargin=10pt,
  innerbottommargin=10pt,
  skipabove=6pt,
  skipbelow=6pt
]{graybox}
 
\hypersetup{colorlinks=true, citecolor=darkblue, linkcolor=darkblue, urlcolor=darkblue}

\title{AI Assistance Reduces Persistence and Hurts Independent Performance}

\author{Grace Liu \\
Carnegie Mellon University \\
{\small \texttt{gliu2@andrew.cmu.edu}} \\
\And
Brian Christian \\
UC Berkeley \\
{\small \texttt{brian.christian@berkeley.edu}} \\
\And
Tsvetomira Dumbalska \\
University of Oxford \\
{\small\texttt{tsvetomira.dumbalska@psy.ox.ac.uk}} \\
\And
Michiel A. Bakker \\
Massachusetts Institute of Technology \\
{\small \texttt{bakker@mit.edu}} \\
\And
Rachit Dubey \\
University of California, Los Angeles \\
{\small \texttt{rdubey@ucla.edu}} \\ 
}

\renewcommand{\normalsize}{\fontsize{10pt}{13pt}\selectfont}
\normalsize
\begin{document}

\ifcolmfinal
\fi

\maketitle

\begin{abstract}
People often optimize for \emph{long-term} goals in collaboration: A mentor or companion doesn't just answer questions, but also scaffolds learning, tracks progress, and prioritizes the other person's growth over immediate results. In contrast, current AI systems are fundamentally \emph{short-sighted} collaborators -- optimized for providing instant and complete responses, without ever saying no (unless for safety reasons). What are the consequences of this dynamic? Here, through a series of randomized controlled trials on human-AI interactions ($N=1,222$), we provide \emph{causal} evidence for two key consequences of AI assistance: reduced persistence and impairment of unassisted performance. Across a variety of tasks, including mathematical reasoning and reading comprehension, we find that although AI assistance improves performance in the short-term, people perform significantly worse without AI and are more likely to give up. These findings are particularly concerning because persistence is foundational to skill acquisition and is one of the strongest predictors of long-term learning. We posit that persistence is reduced because AI conditions people to expect immediate answers, thereby denying them the experience of working through challenges on their own. These results add to the growing literature on cognitive offloading and AI-induced deskilling, and suggest the need for AI model development to prioritize scaffolding long-term competence alongside immediate task completion. 
 \end{abstract}

\begin{center}
  \textbf{Project Page:} \href{https://graliuce.github.io/AI-assistance-reduces-persistence/}{https://graliuce.github.io/AI-assistance-reduces-persistence/} 
\end{center}


\newmdenv[
  topline=true,
  bottomline=true,
  rightline=true,
  leftline=true,
  linewidth=1pt,
  linecolor=darkblue,
  backgroundcolor=gray!4,
  innerleftmargin=14pt,
  innerrightmargin=14pt,
  innertopmargin=12pt,
  innerbottommargin=12pt,
  skipabove=12pt,
  skipbelow=6pt
]{sigbox}

\begin{sigbox}
{\large\bfseries\color{darkblue} Public Significance Statement}\\[3pt]
The rapid rise of AI chatbots promises immediate and effective help with reasoning-intensive tasks such as studying, writing, coding, and brainstorming. But what happens to users' own abilities when the AI is not available? In a series of large-scale human experiments involving arithmetic and reading comprehension, we find that AI assistance improves immediate performance, but it can come at a cognitive cost: after access to AI during problem-solving, people who lost access to the AI performed worse and gave up more frequently than those who never used it. These findings raise urgent questions about the cumulative effects of daily AI use on human persistence and reasoning. We caution that if such effects accumulate with sustained AI use, current AI systems -- optimized only for short-term helpfulness -- risk eroding the very human capabilities they are meant to support. 
\end{sigbox}

\section{Introduction}
\vspace{-0.2cm}
Imagine the following scenario. You are mentoring a student, and they come to you asking you to solve a coding problem. You help them, walking through the solution step by step. They then come back and ask you to solve another problem. And then another. Eventually, you might pause as you recognize that something is going wrong. You realize that your student isn't learning how to code and is simply learning to rely on your help. You subsequently sit them down and talk about the value of persisting through challenges, of practicing new skills, and what it actually means to learn.

This scenario highlights a fundamental aspect of human collaboration. Good collaborators optimize for \emph{long-term} objectives \citep{bratman1992shared, grosz1996collaborative, balcazar2014goals, mattessich2018collaboration}. For example, a mentor encourages independent development by adjusting the type and frequency of help given. In essence, the best collaborators maintain a balance between helping and fostering autonomy; they know when \emph{not} to help \citep{koedinger2007exploring, van2010scaffolding, soderstrom2015learning}. 

Current AI assistants are a stark contrast to this dynamic. They rarely refuse to help, and provide \emph{instant} answers to almost any query, across domains ranging from writing to coding to tutoring \citep{brynjolfsson2025generative, buccinca2024towards, oecd2026digital, shapira2026rlhf}. In this sense, AI systems are fundamentally \emph{short-term} collaborators: extraordinarily helpful in the moment, but indifferent to what that help does to the person receiving it over time.

In this paper, we investigate the consequences of this misalignment between short-term and long-term human-AI collaboration goals. We conduct a series of randomized experiments across $1,222$ participants, spanning mathematical reasoning (Experiments 1 and 2) and reading comprehension (Experiment 3), to provide \emph{causal} evidence for how AI assistance impacts subsequent independent problem-solving capacity.

To preview our findings, we find that AI assistance \textbf{impairs independent performance and reduces persistence}. Although AI assistance improves performance initially, people's performance drops sharply once AI is removed. More strikingly, participants in the AI condition also give up more frequently on the tasks compared to control participants. This pattern is robust and replicates across domains. These effects emerge after a 10--15 minute session, raising the question of what prolonged AI use does to human capability over time.

Our results are notable on two fronts. First, while concern about AI-induced deskilling has grown, prior evidence has been largely correlational \citep{budzyn2025endoscopist, gerlich2025ai} or limited to small samples \citep{kosmyna2025your, shen2026ai}. Recent RCT-based work has begun to establish causal evidence: \cite{bastani2025generative} found that GPT-4 assistance during math practice degraded later unaided performance in high school students, though a scaffolded tutor condition partially mitigated this harm; \cite{barcaui2025chatgpt} similarly showed lasting performance decrements after AI-assisted coursework in undergraduates. Our work builds on and extends this emerging causal literature through a series of randomized controlled trials spanning multiple task domains and population samples beyond educational setting. Second, \textbf{we demonstrate how AI can result in loss of motivation and persistence}. A rich body of literature in cognitive science and education has shown that the capacity to regulate effort and persist through difficulty is foundational to effective learning, and is among the strongest predictors of long-term academic achievement, workforce adaptability, and resilience \citep{metcalfe1999hot, duckworth2007grit, maddux200931, metcalfe2009metacognitive,  andersson2011role, bjork2011making, kapur2014productive, guiso2016long, mooradian2016perspiration}. Our results suggest that AI assistance erodes precisely these capacities. \textbf{People do not merely become worse at tasks, but they also stop trying -- particularly when they use AI to obtain direct answers.} We urge caution in interpreting these results as broad evidence that AI lowers persistence and independent ability for all tasks; nonetheless, at minimum, our findings suggest that AI assistance displaces the very practice opportunities through which people develop lasting competence, a concern echoed in recent perspectives on AI use in medical training~\citep{ke2026ai}.

Yet these findings need not be cause for pessimism. Rather, they point toward a clear design imperative: AI systems should optimize for long-term human capability and autonomy, a goal that cannot be achieved by surface-level interventions \citep{collins2024building, sucholutsky2025using}. This requires rethinking how AI systems are built to collaborate with humans, and just as the best human collaborators know when not to help, so too should AI.

\section{Experiment 1: AI impairs unassisted performance and persistence}

To investigate the causal impact of AI assistance on subsequent problem-solving capacity, we conducted a randomized controlled experiment ($N=354$) on fraction-solving tasks (see also Fig.~\ref{fig:exp_overview}a in Appendix~\ref{App:exp1}). We chose fraction problems as our task for several reasons: their difficulty is easy to experimentally manipulate, they are a foundational mathematical skill \citep{braithwaite2021putting}, struggles with fractions predict later difficulties in mathematics \citep{siegler2013developmental}, and critically, performance on fraction problems improves with practice.

\subsection{Method}
We recruited $354$ US-based participants from the online research platform Prolific and paid them \$2.60 for participation (our study took approximately 13 minutes to complete). In the experiment, participants were given a series of 15 fraction problems to solve of varying difficulty. Participants were explicitly informed that there was no penalty for providing wrong answers, their payment didn't depend on how many questions they solve correctly, and they were requested to do the task to the best of their abilities.

At the beginning of the experiment, participants were randomly assigned to two conditions -- the AI condition ($N = 191$) or the control condition ($N = 163$). Participants in the AI condition were informed that they would have access to an AI assistant for some of the problems and encouraged to use the AI however they liked, with no penalty for doing so. They were then presented with a series of 12 fraction problems with an AI assistant (GPT-5) available in a sidebar. The AI assistant was pre-prompted with each problem and its solution, allowing participants to receive immediate, accurate answers with minimal effort (if they chose to do so). For example, they could simply type ``answer?'', and receive a solution in return (see Appendix~\ref{App:exp1} for experiment details). 

To measure independent problem-solving capacity, the AI assistant was then removed without warning, and participants were asked to solve 3 additional fraction problems. For these problems, participants were requested not to use AI or other external sources. Importantly, these problems were identical across conditions and served as the primary measure of independent performance. Participants in the control condition were presented with the same 15 problems with no assistance and similarly requested not to use AI or other external sources (see Fig.~\ref{fig:exp_overview}a in Appendix~\ref{App:exp1}).

In both conditions, to enable learning from mistakes, if a participant submitted an incorrect answer, the correct solution was shown on the same screen. Furthermore, in both conditions, participants had the option of skipping a problem by clicking a ``skip'' button. Since participants were explicitly told there was no penalty for wrong answers, choosing to skip reflects a deliberate decision not to engage, making it a clear measure of motivation and persistence, independent of ability. \vspace{-0.3cm}

\begin{figure}[t!]
\centering
\includegraphics[width=0.90\linewidth]{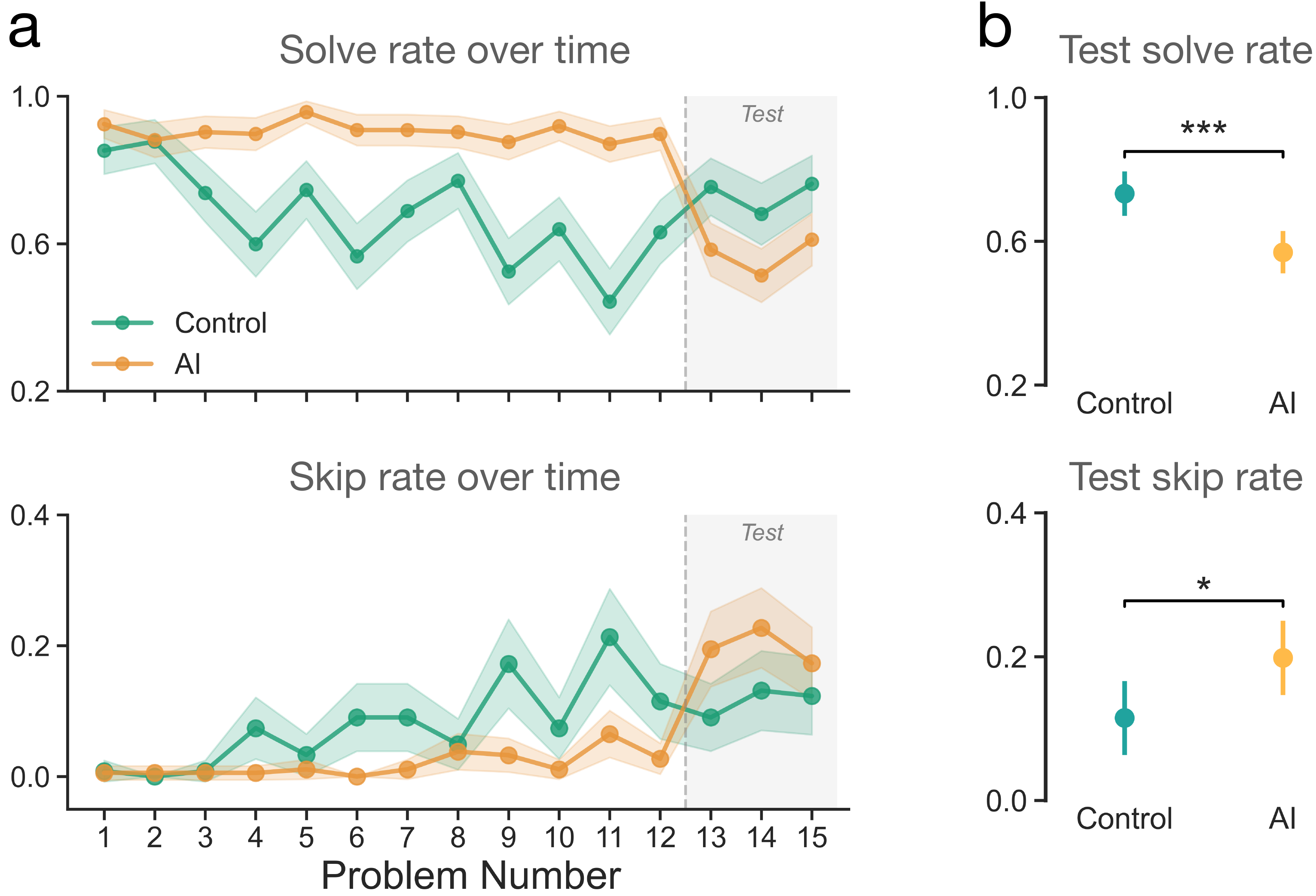}
\caption{
\textbf{AI impairs unassisted performance and persistence.} 
(a) Participants' mean solve rate and skip rate per problem in the order presented, with 95\% confidence intervals (CIs). Dashed gray lines denote the transition between learning and test problems. Problem difficulty increased across the experiment from one-step\protect\footnotemark[1] (problems 1--4) to two-step\protect\footnotemark[2] (problems 5--8) to three-step\protect\footnotemark[3] (problems 9--12). 
(b) Participants' mean test solve rate and skip rate with 95\% CIs across participants. Test metrics are computed by averaging performance over the final three test problems for each participant.
}
\vspace{-0.2cm}
\label{fig:exp1}
\end{figure}

\footnotetext[1]{Example one-step fraction problem: $5/6-1/3$.}
\footnotetext[2]{Example two-step fraction problem: $(7/8 - 1/2) \times 5/6$.}
\footnotetext[3]{Example three-step fraction problem: $(5/6 - 1/4) \times (3/5 + 1/10)$.}
\subsection{Results}

For the results that follow, we used the following criteria for exclusion of data. First, we removed participants who did not pass a simple attention check. We also removed participants who were unable to solve at least 3  of the 12 problems in the learning stage. Given that the first 3 problems were one-step $^{1}$, this exclusion criteria ensured that the remaining participants were capable of solving basic fractions. This led to the exclusion of 49 participants, leaving a final sample of 305 participants ($N = 185$ in the AI condition and $N = 120$ in the control condition).

Figure~\ref{fig:exp1}a shows how participants' solve rate and skip rate changed over time during the experiment. The solve rate measures the rate at which the participants correctly solved the problem, and the skip rate measures the rate at which the participants chose to skip the problem rather than inputting an answer. We observe that with AI assistance (i.e., the first 12 problems), participants in the AI condition were significantly more likely to solve problems correctly and less likely to give up compared to the control participants. However, when the AI was removed, their performance dropped and they were also less likely to persist with problems (as seen by the increased skip rate).

This is further quantified in Figure~\ref{fig:exp1}b, which shows the mean solve and skip rate over the 3 test problems for participants in the two conditions. Participants in the AI condition had a lower solve rate (mean $0.57$, s.d. $0.41$) than participants in the control condition (mean $0.74$, s.d. $0.34$; $t(303) = -3.87$, $P < 0.001$; Cohen's $d = -0.45$, 95\% CI $[-0.69,-0.22]$). Additionally, participants in the AI condition had a higher skip rate (mean $0.20$, s.d. $0.36$) than participants in the control condition (mean $0.11$, s.d. $0.28$; $t(303) = 2.34$, $P = 0.020$; Cohen's $d = 0.27$, 95\% CI $[0.04,0.51]$). The lower solve rate of participants in the AI condition indicates that AI assistance impairs unassisted task performance. Crucially, the higher skip rate suggests that AI assistance also reduced persistence and motivation. 

We note one potential limitation of these results: our exclusion criteria removed participants unable to solve basic fraction problems but didn't account for participants in the AI condition who were similarly unable yet submitted correct answers via AI. This introduced a potential confound in that the AI condition could have selectively retained lower-ability participants. The difference in sample size by condition suggests a higher attrition rate in the control group, potentially inflating the performance gap. We address this confound directly in Experiment 2 and perform additional attrition analysis for the following experiments in Appendix~\ref{App:attrition}.

\section{Experiment 2: Replicating the results and ruling out confounds}
\vspace{-.8em}
In Experiment 2, we conducted a replication of Experiment 1 with two key methodological improvements. First, we added a pretest  of easy one-step fraction problems and used pretest performance for exclusions, rather than in-experiment performance, addressing the skill-level confound described above. Second, we equipped control participants with a sidebar displaying pretest solutions -- information already seen, since solutions were shown after each pretest problem in both conditions -- to eliminate the interface asymmetry introduced by the AI sidebar being present and then suddenly removed (see Fig.~\ref{fig:exp_overview}b in Appendix~\ref{App:exp1}).
\vspace{-.8em}
\subsection{Method}
We recruited 667 US-based participants from the online research platform Prolific and paid them \$3.40 for participation (our study took approximately 15 minutes to complete). At the beginning of the experiment, participants were randomly assigned to two conditions -- the AI condition ($N = 349$) or the control condition ($N = 318$). The experiment consisted of a pretest  phase, a learning phase, and a test phase. For the pretest phase, participants in both conditions were given 3 one-step fraction problems without assistance. After each pretest problem, the correct solution was shown. 
 
After the pretest, the instructions, interface, and AI assistant setup were identical to those in Experiment 1. Participants in the AI condition were presented with a series of 11 fraction problems (Appendix~\ref{App:exp2_problems}), with an AI assistant (GPT-5) available in a sidebar. The AI assistant was then removed, and participants were asked to solve 3 additional fraction problems. Participants in the control condition were presented the same 14 problems without AI assistance. To ensure both conditions experienced the same interface change between the learning and test phases, control participants had access to a sidebar containing the three worked solutions to the pretest  problems, which was removed for the final 3 test problems. Note that this sidebar did not provide new information, since participants in both conditions had already seen these solutions during the pretest  (see Appendix~\ref{App:sidebar}). Lastly, as before, participants had the option of skipping a problem by clicking a ``skip'' button. 

\begin{figure}[t!]
\centering
\includegraphics[width=0.90\linewidth]{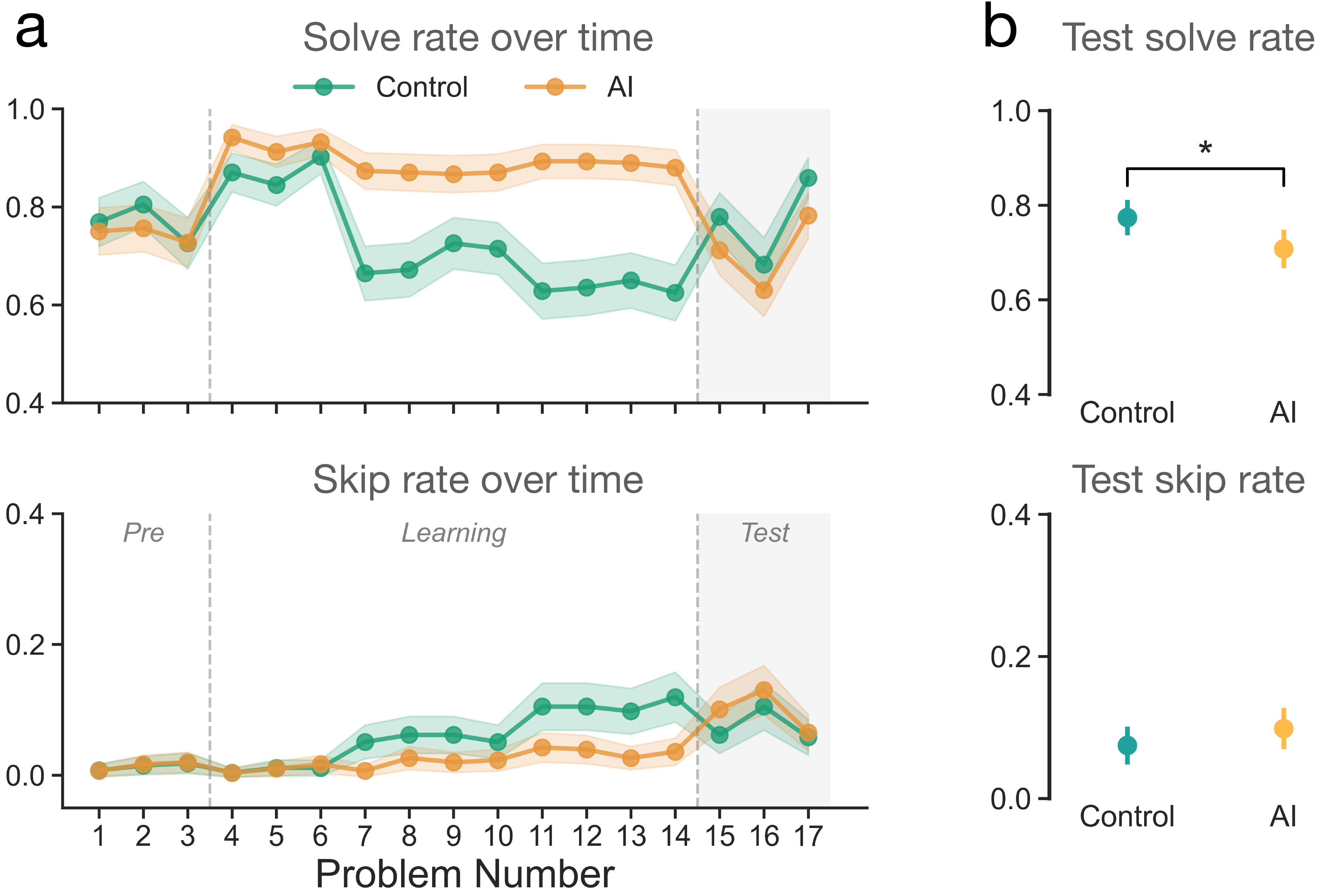}
\caption{\textbf{Replication of results in Experiment 2.} 
(a) Participants' mean solve rate and skip rate per problem in the order presented with 95\% CIs. Problems increased in difficulty from one-step (problems 4-6) to two-step (problems 7-10) to three-step (problems 11-14) problems. (b) Participants' mean test solve rate and test skip rate with 95\% CIs.}
\label{fig:exp2}
\vspace{-0.3cm}
\end{figure} 
\vspace{-0.2cm}
\subsection{Results}
For the results that follow, we first removed participants who did not pass a simple attention check (a 10-point likert-style survey question that asked the participant to select option 10). We then removed participants who did not solve at least 1 out of 3 one-step fraction problems in the pretest. This led to the exclusion of 82 participants, leaving a final sample of 585 participants ($N = 308$ in the AI condition and $N = 277$ in the control condition). We observed similar exclusion rates for the AI-assisted  ($11.7\%$) and control participants ($12.9\%$), suggesting that the skill-level confound from Experiment 1 had been resolved (see also Figure~\ref{fig:attrition_analysis} in Appendix~\ref{App:attrition} for attrition analysis). 

Replicating Experiment 1, AI assistance improved performance during the learning phase, but solve rates dropped and skip rates increased once the AI was removed  (Figure~\ref{fig:exp2}a). Figure~\ref{fig:exp2}b further shows the mean solve rate and skip rate over the final 3 test problems for participants across both conditions. Participants in the AI condition had a lower solve rate (mean $0.71$, s.d. $0.36$) than participants in the control condition (mean $0.77$, s.d. $0.32$; $t(583) = -2.36$, $P = 0.019$; Cohen's $d = -0.20$, 95\% CI $[-0.36,-0.03]$). Participants in the AI condition also exhibited a higher skip rate (mean $0.10$, s.d. $0.26$) than participants in the control condition (mean $0.07$, s.d. $0.23$; $t(583) = 1.18$, $P = 0.239$; Cohen's $d = 0.10$, 95\% CI $[-0.07,0.26]$), but the result was not significant. Although the skip rate difference did not reach significance in the aggregate analysis, this is possibly due to heterogeneity in how participants used the AI. We examine these patterns in the next section.

\subsection{Persistence costs are concentrated among those who obtain direct solutions}

\begin{figure}[t!]
\centering
\includegraphics[width=0.95\linewidth]{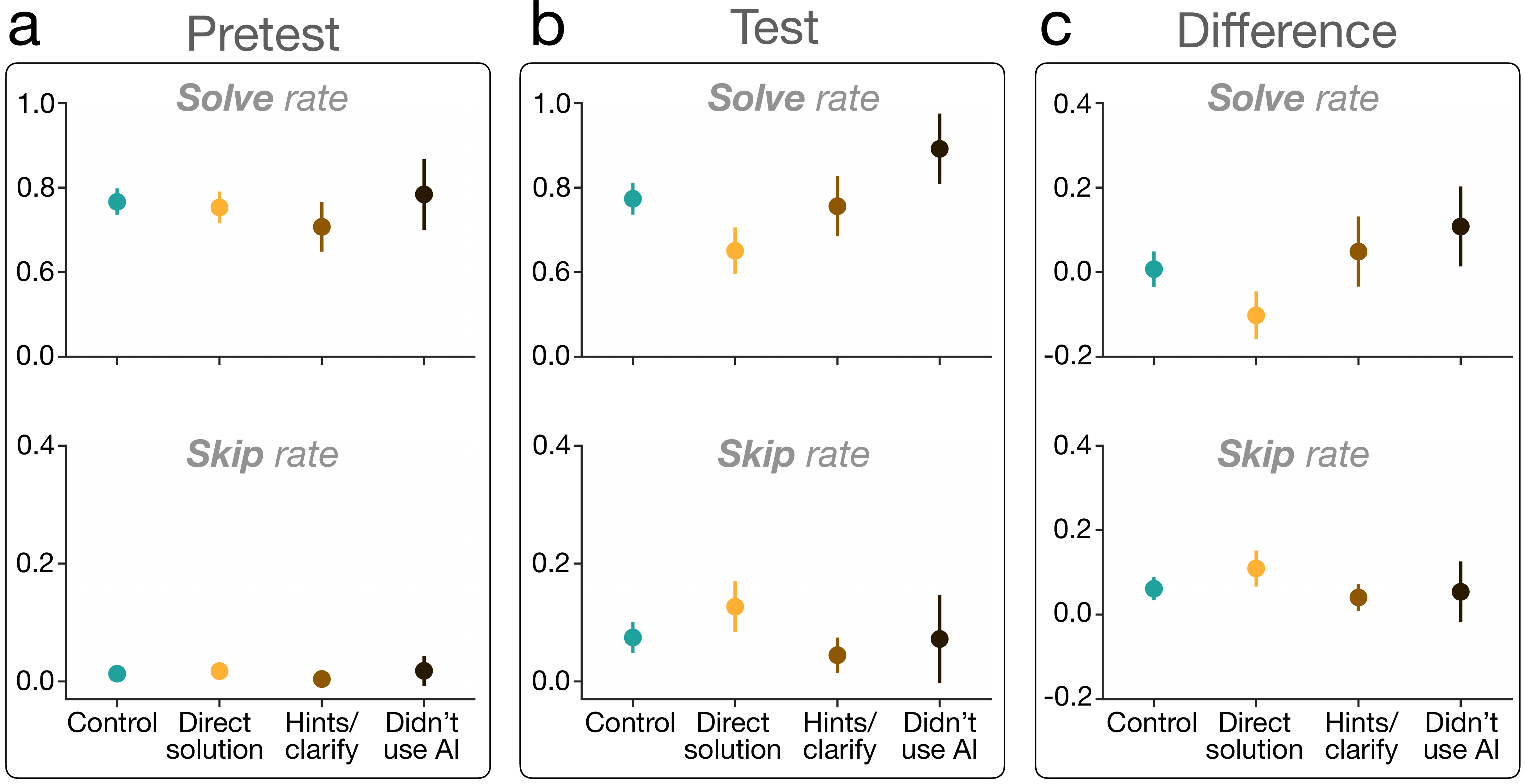}
\caption{\textbf{Performance and persistence declines are concentrated among participants who obtained direct solutions from AI.} (a) AI usage groups show no significant differences in solve rate or skip rate at pretest (one-way ANOVA), suggesting comparable initial skill and motivation levels. (b) Groups differ significantly at test (one-way ANOVA): participants who used AI for direct answers show the lowest solve rate and highest skip rate at test-time. (c) Participants who used AI for direct answers show decline in performance (solve rate) and increased disengagement (skip rate) relative to their own pretest performance. Other groups show similar or improved performance relative to their pretest performance.} 
\label{fig:usage_analysis}
\vspace{-0.5cm}
\end{figure}   

In this section, we analyze the relationship between AI usage patterns and subsequent performance and persistence. Our analysis suggests that the persistence effect is concentrated among participants who used the AI to obtain direct solutions. These results focus on the AI-assisted subset of our RCT data and are cross-sectional in nature; therefore, the findings below are not necessarily causal.

At the end of Experiment 2, we asked participants in the AI condition to self-report how they used the AI assistant during the task (using a multiple choice question). We found that the majority of participants (61\%, $N=189$) in the AI condition self-report that they used the AI primarily to get answers directly. Others reported that they used the AI to get hints or clarifications (27\%, $N=82$), and some participants reported no AI usage (12\%, $N=37$). We validate self-reported AI usage with chatlog records. Out of the 37 participants that reported no AI usage, 83.8\% did not message the AI at all and 94.1\% messaged less than three times.

Figures~\ref{fig:usage_analysis}a and ~\ref{fig:usage_analysis}b show the solve rate and skip rate of the participants grouped by AI usage type for the pretest and the final test. For comparison, we also include data from the control participants. Using a one-way ANOVA test, we found that there was no significant difference between the AI usage groups as well as the control group for the pretest solve rate ($F(3, 581) = 1.32$, $P = 0.266$, $\eta^2 = 0.007$) nor the pretest skip rate ($F(3, 581) = 0.56$, $P = 0.639$, $\eta^2 = 0.003$), implying that participants entered the experiment with similar skill and motivation levels. However, a one-way ANOVA test revealed a significant difference between the groups for both test solve rate ($F(3, 581) = 8.15$, $P < 0.001$, $\eta^2 = 0.040$) and test skip rate ($F(3, 581) = 2.80$, $P = 0.039$, $\eta^2 = 0.014$).

To investigate this difference, we conducted pairwise t-tests between the groups. Participants who used the AI to get answers directly demonstrated lower subsequent task performance (mean $0.65$, s.d. $0.38$) compared to participants in the control group (mean $0.77$, s.d. $0.32$; $t(464) = -3.84$, $P < 0.001$; Cohen's $d = -0.36$, 95\% CI $[-0.55$,-$-0.18]$), participants who used the AI to get hints (mean $0.76$, s.d. $0.32$; $t(269) = -2.25$, $P = 0.025$; Cohen's $d = -0.30$, 95\% CI $[-0.56,-0.04]$), and participants who did not use the AI (mean $0.89$, s.d. $0.25$; $t(224) = -3.73$, $P < 0.001$; Cohen's $d = -0.67$, 95\% CI $[-1.03,-0.31]$). Furthermore, participants who used the AI to get answers directly had a higher skip rate (mean $0.13$, s.d. $0.30$) than participants in the control group (mean $0.07$, s.d. $0.23$; $t(464) = 2.14$, $P = 0.033$; Cohen's $d = 0.20$, 95\% CI $[0.02$, $0.39]$) and participants who used the AI to get hints (mean $0.05$, s.d. $0.14$; $t(269) = -2.36$, $P = 0.019$; Cohen's $d = -0.31$, 95\% CI $[-0.57$, $-0.05]$). Participants who did not use AI had a mean skip rate of $0.07$ (s.d. $0.22$). Thus, when people prompted AI to solve tasks for them, they were less likely to persist with tasks compared to people who didn't use AI or people who used AI to aid understanding.

To conclude this analysis, Fig.~\ref{fig:usage_analysis}c shows how participants' test performance (skip and solve rate) changed relative to their pretest performance. These metrics are computed by taking the average over all participants of their test solve/skip rate minus their pretest solve/skip rate. Participants who used the AI to get answers had a larger decrease in test solve rate from their pretest solve rate (mean $-0.11$, s.d. $0.39$) compared to control participants (mean = $0.01$, s.d. = $0.35$; $t(464) = -3.25$, $P = 0.001$; Cohen's $d = -0.31$, 95\% CI $[-0.49$, $-0.12]$), participants who used the AI to get hints (mean $0.05$, s.d. $0.37$; $t(269) = -3.11$, $P = 0.002$; Cohen's $d = -0.41$, 95\% CI $[-0.67$, $-0.15]$) and participants who did not use the AI (mean $0.11$, s.d. $0.28$; $t(224) = -3.16$, $P = 0.002$; Cohen's $d = -0.57$, 95\% CI $[-0.93$,$-0.21]$). Further, participants who used the AI to get answers directly showed a larger increase in skip rate from pretest to test (mean $0.11$, s.d. $0.30$) compared to control participants (mean $0.06$, s.d. $0.23$; $t(464) = 1.96$, $P = 0.051$; Cohen's $d = 0.18$, 95\% CI [$0.00$,$0.37]$), and participants who used the AI to get hints (mean $0.04$, s.d. $0.14$; $t(269) = 1.99$, $P = 0.047$; Cohen's $d = 0.26$, 95\% CI $[0.00$, $0.52]$). Though the difference relative to control was marginal ($P=0.051$), the direction was consistent to previous results.

These results show how reliance on AI negatively impacts persistence and independent performance. Participants who prompted the AI for answers directly not only performed worse than all other groups at test, but also declined relative to their own pretest performance.

\section{Experiment 3: Convergent evidence from reading comprehension}
Having established that AI impairs unassisted performance and persistence in mathematical problem-solving, we next asked whether this effect generalizes beyond arithmetic. Experiment 3 replicated our design in the domain of reading comprehension -- a task that draws on fundamentally different cognitive skills, namely meaning-making and mental model construction~\citep{pourhosein2016can,kintsch1998comprehension}, and one with clear stakes for academic success~\citep{bastug2014structural}. All problems were sourced from free, online SAT reading practice materials (see Appendix~\ref{App:exp3}). 

\subsection{Method}
We recruited $201$ US-based participants from the online research platform Prolific and paid them \$3.40 for participation (our study took approximately 13 minutes to complete). At the beginning of the experiment, participants were randomly assigned to two conditions -- the AI condition ($N = 104$) or the control condition ($N = 97$). The instructions for this experiment were similar to those in the fraction experiments (see Appendix~\ref{App:exp3} for details), and the experiment consisted of a pretest phase, a learning phase, and a test phase. For the pretest phase, participants in both conditions were given one simple reading comprehension question without assistance. 

Participants in the AI condition were then presented with a series of 5 reading comprehension problems, with an AI assistant (GPT-5) available in a sidebar. The AI assistant was then removed, and participants were asked to solve 3 additional reading comprehension problems. Participants in the control condition were presented all 8 problems without AI assistance. To ensure that participants in both conditions experience a context-shift between the learning phase and testing phase, control participants had access to a sidebar displaying general test-taking tips (see Appendix~\ref{App:reading_sidebar}). The sidebar was taken away for the final 3 problems. As in prior experiments, participants had the option of skipping a problem by clicking a ``skip'' button. We also recorded a problem as ``skipped'' if the participant provided their answer in less than 5 seconds, as it would have been impossible for them to read the texts in this time frame.

\subsection{Results}
\begin{figure}[t!]
\centering
\includegraphics[width=0.90\linewidth]{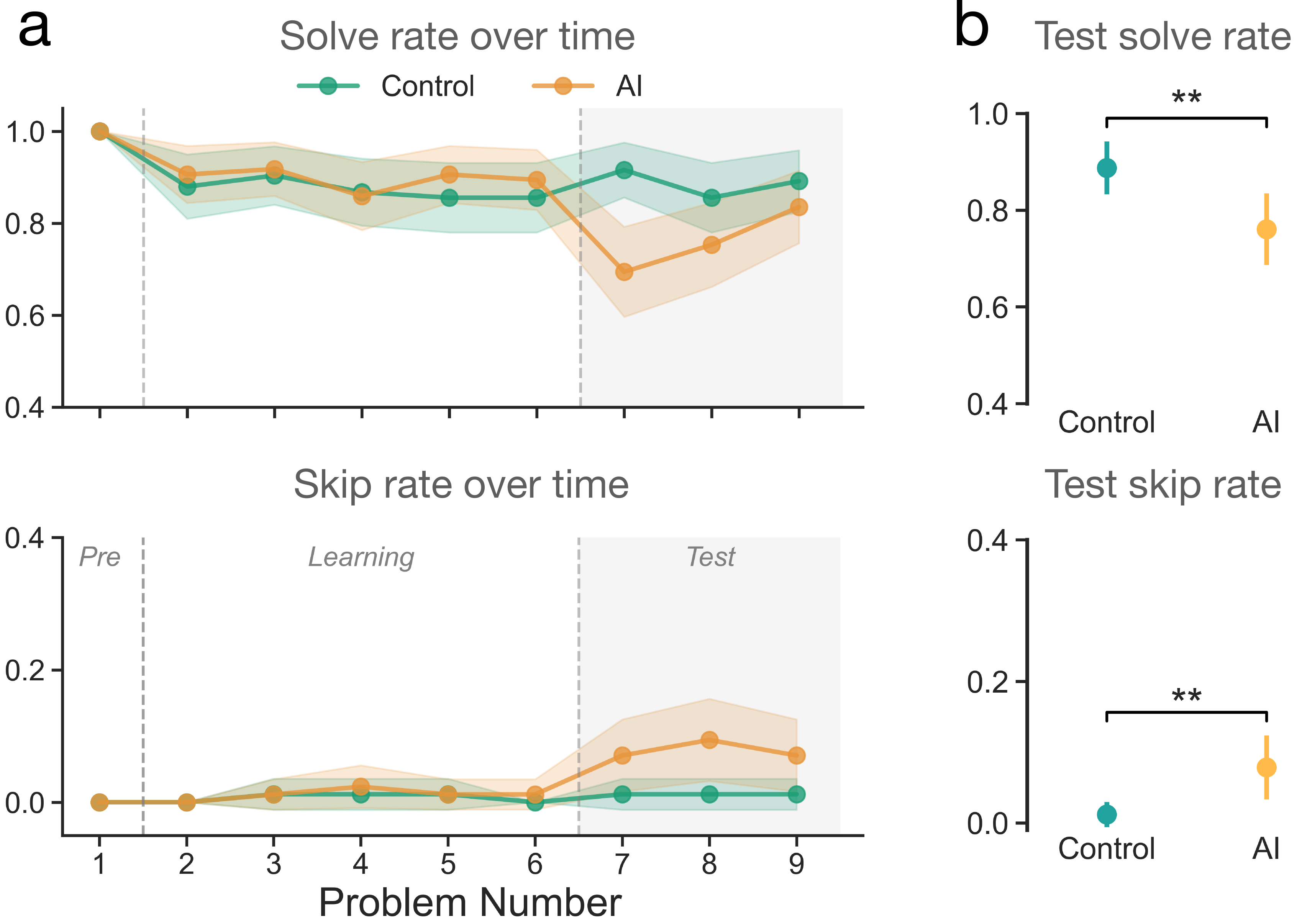}
\caption{\textbf{Reduced performance and persistence in reading comprehension task.}
(a) Participants’ mean solve rate and skip rate per problem in the order presented with 95\% CI. Dashed gray lines denote transition between learning problems and test problems. (b) Participants’ mean test solve rate and test skip rate with 95\% CIs computed across the participants.}
\label{fig:exp3}
\vspace{-0.4cm}
\end{figure}   

For the results that follow, we excluded participants who did not pass a simple attention check and those who did not solve the pretest  problem. This led to the exclusion of 33 participants, leaving a final sample of 168 participants ($N = 85$ in the AI condition and $N = 83$ in the control condition).

As in previous experiments, solve rates dropped and skip rates increased once the AI was removed (Fig.~\ref{fig:exp3}a). Figure~\ref{fig:exp3}b shows the mean solve rate and skip rate over the final 3 test problems for participants in the AI condition vs. participants in the control condition. Participants in the AI condition had a lower solve rate (mean $0.76$, s.d. $0.34$) than participants in the control condition (mean $0.89$, s.d. $0.25$; $t(166) = -2.72$, $P = 0.007$; Cohen's $d = -0.42$, 95\% CI $[-0.73,-0.11]$). Participants in the AI condition also had a higher skip rate (mean $0.08$, s.d. $0.21$) than participants in the control condition (mean $0.01$, s.d. $0.08$; $t(166) = 2.69$, $P = 0.008$; Cohen's $d = 0.42$, 95\% CI $[0.11,0.72]$). The lower solve rate of participants in the AI condition demonstrates that AI assistance impairs unassisted performance for a task that is closely intertwined with critical thinking. Again, the higher skip rate of this group suggests that AI assistance also reduces persistence and motivation in reading tasks. This replication suggests that reduced persistence is not an artifact of mathematical reasoning tasks but a general consequence of AI-assisted problem solving.

\vspace{-0.2cm}
\section{Related Work}
\vspace{-0.2cm}
\textbf{Cognitive offloading.} Cognitive offloading~\citep{risko2016cognitive} is the use of physical action (e.g. writing notes, typing in a calculator, using a search engine) to alter the information processing requirements of a task and reduce cognitive demand ~\citep{goldin2001explaining,martin2005physically,gilbert2020optimal}. However, this improved task performance also risks decline in performance when the cognitive aid is not available~\citep{sparrow2011google,grinschgl2021consequences,richmond2025benefits}.
AI accelerates this dynamic across virtually every reasoning domain~\citep{kosmyna2025your,shen2026ai}, raising concerns of overreliance~\citep{ibrahim2025measuring,kim2025fostering} and deskilling~\citep{budzyn2025endoscopist,shen2026ai}. Existing evidence, however, is largely correlational through surveys or interviews~\citep{gerlich2025ai,lee2025impact}. We address this limita tion by providing causal evidence through a series of RCTs.

\textbf{Human-AI collaboration.} A growing line of work considers how to design human-AI systems that optimize for long-term outcomes rather than short-term preference satisfaction ~\citep{sumers2022talk,collins2024building, sucholutsky2025using, steyvers2025not, zhao2025comparing, cheng2026accommodation}. Short-term optimization often produces misalignment between what AI systems are rewarded for and what users need \citep{christian2026reward}, giving rise to sycophancy~\citep{cheng2025sycophantic, cheng2025social, rathje2025sycophantic}, manipulation~\citep{williams2024targeted}, and deception~\citep{zhou2025emergent}. We argue that cognitive deskilling represents an underappreciated instance of this misalignment: AI systems optimized for immediate helpfulness may undermine the long-term capabilities of the people they aim to assist.

\textbf{Gradual AI risks.} AI safety research has traditionally focused on abrupt, dire events caused by advanced AI systems ~\citep{bucknall2022current,carlsmith2022power, bengio2024managing}. A growing literature instead considers the possibility of AI risks manifesting \emph{gradually} through an incremental series of misalignment in objectives between human preferences and AI-dependent societal systems~\citep{kasirzadeh2025two, kulveit2025gradual}. However, there is limited empirical evidence on how this ``boiling frog'' scenario of incremental AI risks will play out, especially at the individual level \citep{ibrahim2025measuring}. Although related work has explored the potential of AI usage to reduce human agency in interpersonal relationships~\citep{kirk2025neural,sharma2026s}, creative output~\citep{padmakumar2023does, doshi2024generative}, and personal beliefs~\citep{jakesch2023co,williams2024targeted, guingrich2026belief}, no prior work has examined how AI use erodes cognitive processes and undermines autonomous thought. Our work addresses this gap directly, providing cognitive evidence for a ``gradual disempowerment'' hypothesis. Sustained AI use risks eroding essential cognitive and motivational capacities, and as a consequence, displacing human participation over time. 

\vspace{-0.3cm}
\section{Conclusion}
\vspace{-0.3cm}

Human cognition has always been shaped by external tools, from calculators to internet to GPS navigation. Current AI systems, however, represent a new kind of cognitive scaffold: one that solves anything, rarely refuses to help, and delivers answers instantly.  Here, we show that  AI interaction can result in significant impairments in independent performance and persistence -- capacities that are foundational to life-long learning. Although our study only explores brief exposure to AI, it is important for the field to study the sustained impacts of AI use on our minds. If our findings hold true for longer periods of time, then the cumulative effects of daily AI use over months or years may be profound and difficult to reverse.

Two mechanisms may explain the observed decline in persistence. First, when AI routinely completes tasks in seconds, the reference point for how long a task \emph{should} take can shift -- and as a consequence, unaided work starts to feel counterfactually more effortful, a process structurally analogous to hedonic adaptation~\citep{brickman1971hedonic, brickman1978lottery, frederick199916}. Crucially, this mechanism is self-reinforcing: each act of offloading shifts the reference point, increases the subjective cost of unaided effort, and makes future offloading more attractive. Second, AI removes the productive struggle through which people develop not only accurate knowledge but accurate \emph{self}-knowledge. Without opportunities to work independently, people never learn what they are capable of, undermining the metacognitive calibration that sustains persistence \citep{yeung2012metacognition, fleming2017self, dubey2021aha, elizondo2024self}. Together, these mechanisms predict broader metacognitive decay beyond persistence alone -- a direction future work should examine in naturalistic, longitudinal settings.

Our results carry important policy implications. The tasks investigated here, such as fraction arithmetic and reading comprehension, may seem delegable to tools like calculators, but conceptual mastery of these skills is a \emph{developmental} prerequisite. Without these skills, higher-order competencies like algebra or critical reasoning remain inaccessible. If sustained AI use erodes the motivation and persistence that drive long-term learning, these effects will accumulate over years, and by the time they are visible, they will be difficult to reverse. This is analogous to the ``boiling frog'' effect, where each incremental act feels costless, until the cumulative effect becomes overwhelming to address \citep{moore2019rapidly, kasirzadeh2025two, liu2025binary}. These risks are also not equally distributed: students with fewer academic resources may be most vulnerable. While user-facing interventions (e.g., Socratic AI, reduced use time, etc.), might help at the margins, we believe that such solutions will only serve as ``band-aids'' and will not resolve the deeper issue, since AI offers a temptation to offload at scale. Mitigating these risks requires rethinking how AI systems collaborate with people long-term and broadening  objectives beyond short-term user satisfaction \citep{zhi2025beyond, kirk2025human} toward an ethos of empowerment and care \citep{kleiman2024computational, christian2025computational}. We hope that our work inspires the field to think about optimizing not just what people can do \emph{with} AI, but what they can do \emph{without} it. 

\textbf{Limitations and future directions.} It is important to note that our experiments captured only brief AI exposure and whether these effects accumulate or attenuate with prolonged daily use remains an open question. Second, we cannot speak to the durability of these effects: participants were tested immediately after AI removal, and whether the impairments persist hours (or days) later is unknown. Third, our AI was maximally helpful, providing complete answers on demand. Whether a Socratic AI could preserve assistance benefits while avoiding persistence costs is an important open question, and one that the hint-usage pattern in Experiment 2 tentatively suggests is worth pursuing. Addressing these limitations, particularly the first two, requires longitudinal experiments that track the cumulative effects of AI use on independent performance and persistence over multiple sessions.


\section*{Acknowledgments}
GL acknowledges support by the National Science Foundation Graduate Research Fellowship under Grant No. DGE2140739. Any opinions, findings and conclusions or recommendations expressed in this material are those of the author(s) and do not necessarily reflect the views of the National Science Foundation. MB acknowledges support from Galen Hines-Pierce.

\section*{Ethics Statement}
All experiments reported in the paper were approved by the institutional IRB at University of California, Los Angeles (IRB-25-2012). Participants provided informed consent prior to the experiment and were compensated for their time (at a rate of \$15 per hour). They were informed that they could withdraw at any time without penalty. No personally identifiable information was collected. The tasks involved short, cognitive exercises and posed no risk of harm to participants.

\appendix

\begin{figure}[t!]
\centering
\includegraphics[width=0.85\linewidth]{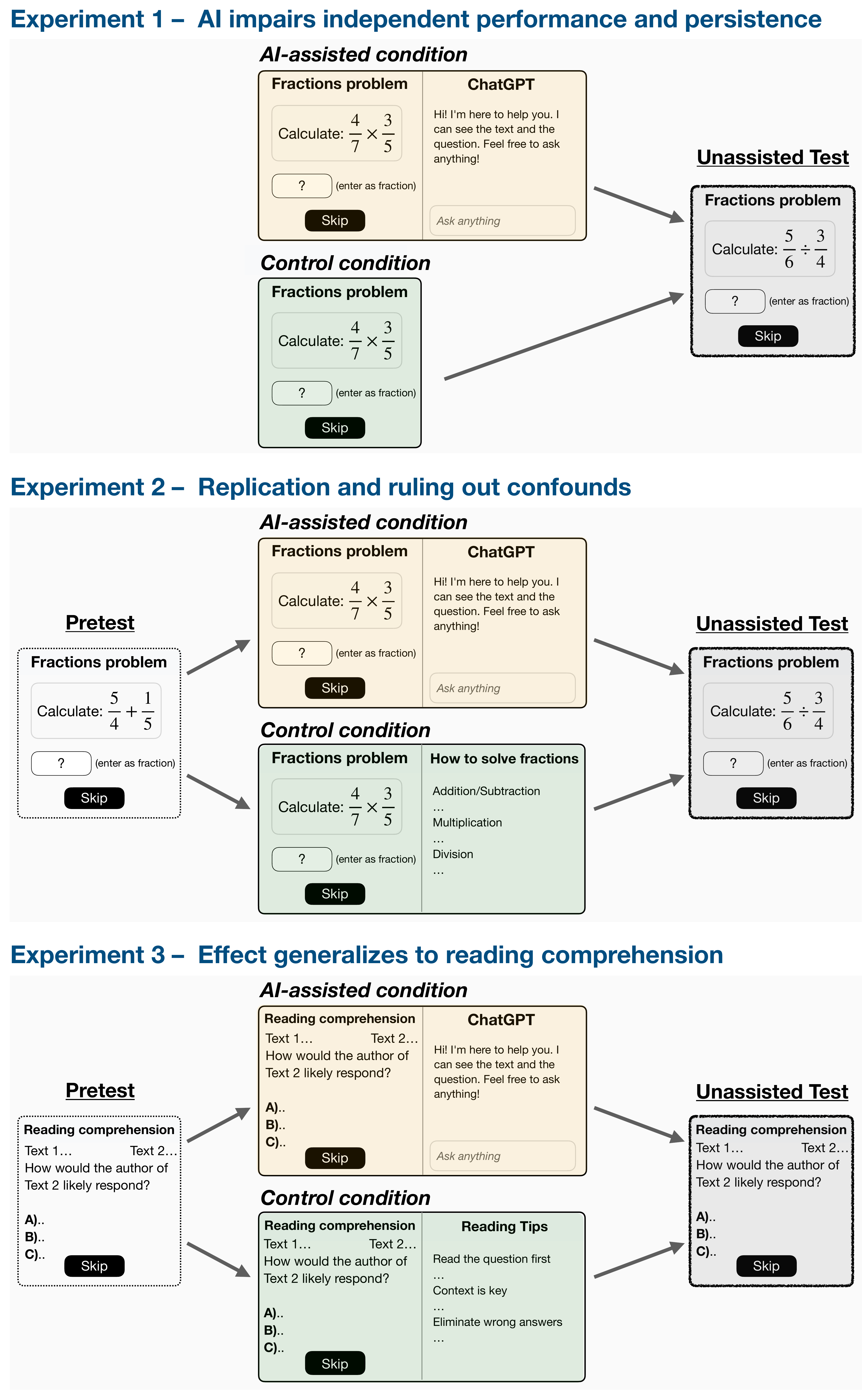}
\caption{\textbf{Overview of Experiments.} Top panel: Experiment 1 shows that AI impairs independent performance and persistence. Middle panel: Experiment 2 replicates the effect on a larger population and address potential confounds. Bottom: Experiment 3 replicates the effect on a reading comprehension task} 
\label{fig:exp_overview}
\end{figure}   

\newpage
\section{Experiment 1 Details}\label{App:exp1}
\subsection{Participant Instructions}\label{App:exp1_instructions}

\paragraph{Control condition:}

\begin{graybox}
\textit{Welcome! Thank you for participating in this study about how different
learning interventions can improve learning outcomes.}

\medskip
\textit{In this study, you will be solving fraction problems. These include both
direct calculations and word problems involving fractions.}

\medskip
\textit{You have been assigned to the self-learning condition. In this condition,
you will work through the problems on your own.}

\medskip
\textit{Please don't use AI tools or Google to solve the problems.}

\medskip
\textit{Enter answers as fractions (e.g., 3/4) or mixed numbers (e.g., 2\,1/2).}

\medskip
\textit{Important: Your payment does NOT depend on how many problems you answer
correctly.}
\end{graybox}

\paragraph{AI-assisted condition:}

\begin{graybox}
\textit{Welcome! Thank you for participating in this study about how different
learning interventions can improve learning outcomes.}

\medskip
\textit{In this study, you will be solving fraction problems. These include both
direct calculations and word problems involving fractions.}

\medskip
\textit{You have been assigned to the AI-assisted learning condition. In this
condition, you will have access to an AI assistant (ChatGPT) to help you solve
problems. Please try to use the AI to help you work through the problems.}

\medskip
\textit{For some problems, ChatGPT won't be available. When ChatGPT isn't available,
please don't use other AI tools or Google.}

\medskip
\textit{Enter answers as fractions (e.g., 3/4) or mixed numbers (e.g., 2\,1/2).}

\medskip
\textit{Important: Your payment does NOT depend on how many problems you answer
correctly.}
\end{graybox}

\newpage
\subsection{Problems}~\label{App:exp1_problems}

All problems required participants to enter their answer as a fraction or mixed
number; decimal answers were not accepted. Problems were presented in fixed order (progressively harder). The AI chat assistant was available to participants in the AI-assisted condition during the main phase but not the final phase.

\paragraph{Main Problems (12 problems):}

\begin{graybox}
\begin{enumerate}
  \item Calculate: $\frac{3}{5} + \frac{1}{4}$
  \item Calculate: $\frac{5}{6} - \frac{1}{3}$
  \item Calculate: $\frac{2}{3} \times \frac{4}{5}$
  \item Calculate: $\frac{3}{4} \div \frac{2}{3}$
  \item Calculate: $\left(\frac{2}{3} + \frac{1}{4}\right) \times \frac{3}{5}$
  \item Calculate: $\left(\frac{5}{6} - \frac{1}{3}\right) \div \frac{2}{5}$
  \item Calculate: $\left(\frac{3}{4} + \frac{2}{5}\right) \times \frac{4}{7}$
  \item Calculate: $\left(\frac{7}{8} - \frac{1}{2}\right) \times \frac{5}{6}$
  \item Calculate: $\left(\left(\frac{1}{2} + \frac{1}{3}\right) \times \frac{3}{4}\right) \div \frac{2}{5}$
  \item Calculate: $\left(\frac{3}{4} - \frac{1}{2}\right) \times \left(\frac{1}{3} + \frac{1}{6}\right)$
  \item Calculate: $\left(\left(\frac{2}{3} + \frac{1}{4}\right) \times \frac{6}{7}\right) \div \frac{3}{5}$
  \item Calculate: $\left(\frac{1}{2} \times \frac{3}{5}\right) + \left(\frac{2}{3} \times \frac{3}{4}\right)$
\end{enumerate}
\end{graybox}

\paragraph{Final Problems (Test Phase, 3 problems):}

Neither the AI chat assistant nor any reference materials were available during this
phase for either condition.

\begin{graybox}
\begin{enumerate}
  \item Calculate: $\left(\frac{2}{3} + \frac{1}{4}\right) \times \frac{3}{5}$
  \item Calculate: $\left(\frac{3}{4} - \frac{1}{2}\right) \times \left(\frac{2}{3} + \frac{1}{4}\right)$
  \item Calculate: $\frac{3}{4} \div \frac{1}{2}$
\end{enumerate}
\end{graybox}

\subsection{AI-Assisted Condition: Chat Assistant Greeting}

Participants in the AI-assisted condition had access to a chat interface (described
to participants as ChatGPT) throughout the main phase. The assistant opened each
problem with the following greeting:

\begin{graybox}
\textit{``Hi! I'm here to help you solve the problem. I can already see the problem
you need to solve. Feel free to ask anything!''}
\end{graybox}

The assistant had access to the text of the currently displayed problem and could
respond to any free-text queries from the participant.

\newpage
\section{Experiment 2 Details}\label{App:exp2}

\subsection{Participant Instructions}\label{App:exp2_instructions}

\paragraph{General instructions (all participants):}

After providing informed consent, all participants received the following opening
instructions:

\begin{graybox}
\textit{Welcome! Thank you for participating in this study about how different
learning interventions can improve learning outcomes.}

\medskip
\textit{In this study, you will be solving fraction problems. These include both
direct calculations and word problems involving fractions.}

\medskip
\textit{Enter answers as fractions (e.g., 3/4) or mixed numbers (e.g., 2\,1/2).}

\medskip
\textit{Let's do a practice round so that you can get familiar with the interface.}
\end{graybox}

Following the pretest round, participants were shown one of two
instruction screens depending on their randomly assigned condition.

\paragraph{Control condition:}

\begin{graybox}
\textit{You have been assigned to the self-learning condition. In this condition, we
want you to work through the fraction problems on your own. We highly encourage you
to use paper and pen, and take your time to work through the solutions.}

\medskip
\textit{You will have access to a reference panel with tips on how to solve
fractions.}

\medskip
\textit{Please don't use AI tools, Calculator, or Google to solve the problems.}

\medskip
\textit{Important: Your payment does NOT depend on how many problems you answer
correctly.}
\end{graybox}

\paragraph{AI-assisted condition:}

\begin{graybox}
\textit{You have been assigned to the AI-assisted learning condition. In this
condition, you will have access to an AI assistant (ChatGPT) to help you solve
problems. We encourage you to use the AI to help you work through the problems. You
can use the AI however you like---feel free to ask it for hints, answers, or
solutions.}

\medskip
\textit{For some problems, ChatGPT won't be available. When ChatGPT isn't available,
please don't use other AI tools, Calculator, or Google.}

\medskip
\textit{Important: Your payment does NOT depend on how many problems you answer
correctly.}
\end{graybox}

\newpage
\subsection{Problems}~\label{App:exp2_problems}

Problems were presented in three stages: a pretest (practice) phase, a main
experimental phase, and a final (test) phase. All problems required participants to
enter their answer as a fraction or mixed number (e.g., \texttt{3/4} or
\texttt{2\,1/2}); decimal answers were not accepted.

\paragraph{Pretest Problems:}

Three practice problems were presented to all participants. Neither the AI chat assistant nor the reference hints panel was
available during this phase.

\begin{graybox}
\begin{enumerate}
  \item Calculate: $\frac{3}{5} + \frac{1}{4}$
  \item Calculate: $\frac{2}{3} \times \frac{1}{2}$
  \item Calculate: $\frac{1}{2} \div \frac{1}{4}$
\end{enumerate}
\end{graybox}
\paragraph{Main Problems (11 problems):}
The problems were arranged in three difficulty blocks. Within each block the order
was randomised across participants; block order was fixed (easy $\to$ medium $\to$
hard). The AI chat assistant was available to participants in the AI-assisted
condition; the reference hints panel was available to participants in the control
condition.
\begin{graybox}
\begin{enumerate}
  \item Calculate: $\frac{1}{2} \div \frac{1}{3}$
  \item Calculate: $\frac{1}{3} \div \frac{2}{4}$
  \item Calculate: $\frac{5}{6} - \frac{1}{3}$
  \item Calculate: $\left(\frac{2}{3} + \frac{1}{4}\right) \times \frac{3}{5}$
  \item Calculate: $\left(\frac{5}{6} - \frac{1}{3}\right) \div \frac{2}{5}$
  \item Calculate: $\left(\frac{3}{4} + \frac{2}{5}\right) \times \frac{4}{7}$
  \item Calculate: $\left(\frac{7}{8} - \frac{1}{2}\right) \times \frac{5}{6}$
  \item Calculate: $\left(\left(\frac{1}{2} + \frac{1}{3}\right) \times \frac{3}{4}\right) \div \frac{2}{5}$
  \item Calculate: $\left(\left(\frac{2}{3} + \frac{1}{4}\right) \times \frac{6}{7}\right) \div \frac{3}{5}$
  \item Calculate: $\left(\frac{1}{2} \times \frac{3}{5}\right) + \left(\frac{2}{3} \times \frac{3}{4}\right)$
  \item Calculate: $\left(\frac{5}{6} - \frac{1}{4}\right) \times \left(\frac{3}{5} + \frac{1}{10}\right)$
\end{enumerate}
\end{graybox}

\paragraph{Final Problems (Test Phase):}

Three problems were presented at the end of the study. Neither the AI chat assistant
nor the reference hints panel was available during this phase for either condition.
These problems were presented in fixed order.

\begin{graybox}
\begin{enumerate}
  \item Calculate: $\left(\frac{2}{3} + \frac{1}{4}\right) \times \frac{3}{5}$
  \item Calculate: $\left(\frac{3}{4} - \frac{1}{2}\right) \times \left(\frac{2}{3} + \frac{1}{4}\right)$
  \item Calculate: $\frac{3}{4} \div \frac{1}{2}$
\end{enumerate}
\end{graybox}

\subsection{Scoring}\label{App:exp2_scoring}

Responses were scored automatically - both reduced and unreduced fractions that matched the correct answer value were marked correct. Participants were instructed to answer in fraction format - we made this choice to dissuade the use of calculators. If a participant answered in decimal format, their answer was marked incorrect automatically. Decimal responses were rare, consisting of 0.14\% of all responses. If a participant mistyped their response (i.e. entered a wrong number accidentally), then it was automatically scored incorrect. Skips were counted as wrong answers.

\subsection{Reference Materials Available to Participants}~\label{App:sidebar}
\vspace{-3em}
\paragraph{Control condition: Reference hints panel.}

Participants in the control condition had access to a fixed sidebar
panel titled \textit{``How to Solve Fractions''} throughout the main experimental
phase. The panel contained three solutions to the pretest problems. These solutions did not provide the control participants with new information because participants in both conditions were shown these solutions after each pretest problem.

\textit{Addition \& Subtraction:}

\begin{graybox}
To add or subtract fractions, first find a \textbf{common denominator}. Convert each
fraction so they have the same denominator, then add or subtract the numerators.

\medskip
\textit{Example:} $\frac{3}{5} + \frac{1}{4} = \frac{12}{20} + \frac{5}{20} = \frac{17}{20}$
\end{graybox}

\textit{Multiplication:}

\begin{graybox}
Multiply the numerators together and multiply the denominators together.

\medskip
\textit{Example:} $\frac{2}{3} \times \frac{1}{2} = \frac{2}{6} = \frac{1}{3}$
\end{graybox}

\textit{Division:}

\begin{graybox}
To divide fractions, multiply by the \textbf{reciprocal} (flip the second fraction).

\medskip
\textit{Example:} $\frac{1}{2} \div \frac{1}{4} = \frac{1}{2} \times \frac{4}{1} = \frac{4}{2} = 2$
\end{graybox}

\paragraph{AI-assisted condition: Chat assistant greeting.}

Participants in the AI-assisted condition had access to a chat interface (described
to participants as ChatGPT) throughout the main experimental phase. The assistant
opened each problem with the following greeting:

\begin{graybox}
\textit{``Hi! I'm here to help you solve the problem. I can already see the problem
you need to solve. Feel free to ask anything!''}
\end{graybox}

The assistant had access to the text of the currently displayed problem and could
respond to any free-text queries from the participant.

\newpage
\section{Experiment 3 Details}\label{App:exp3}

\subsection{Participant Instructions}\label{App:exp3_instructions}

\paragraph{General instructions (all participants):}

After providing informed consent, all participants received the following opening
instructions:

\begin{graybox}
\textit{Welcome! Thank you for participating in this study about how different
learning interventions can improve learning outcomes.}

\medskip
\textit{In this study, you will be solving reading comprehension problems. You will
be presented with a short text and asked a question about it.}

\medskip
\textit{Select the best option from the multiple-choice list.}

\medskip
\textit{Let's do a practice round so that you can get familiar with the interface.}
\end{graybox}

Following the pretest round, participants were shown one of two
instruction screens depending on their randomly assigned condition.

\paragraph{Control condition:}

\begin{graybox}
\textit{You have been randomly assigned to the self-learning condition. In this
condition, we want you to work through the reading comprehension problems on your
own. You will have access to a reference panel with tips on how to solve reading
comprehension questions. We highly encourage you to take your time to work through
the solutions.}

\medskip
\textit{For some problems, the reference panel will not be available.}

\medskip
\textit{Please don't use AI tools or Google to solve the problems.}

\medskip
\textit{Important: Your payment does NOT depend on how many problems you answer
correctly.}
\end{graybox}

\paragraph{AI-assisted condition:}

\begin{graybox}
\textit{You have been randomly assigned to the AI-assisted learning condition. In
this condition, you will have access to an AI assistant (ChatGPT) to help you solve
problems. We encourage you to use the AI to help you work through the problems. You
can use the AI however you like -- feel free to ask it for hints, answers, or
solutions.}

\medskip
\textit{For some problems, ChatGPT won't be available. When ChatGPT isn't available,
please don't use other AI tools or Google.}

\medskip
\textit{Important: Your payment does NOT depend on how many problems you answer
correctly.}
\end{graybox}

\newpage
\subsection{Problems}\label{App:exp3_problems}

Each problem presented participants with two short texts taking positions on
a topic, followed by a four-option multiple-choice question. The questions were sourced from free, online SAT reading practice questions from the Manhattan Review\footnote{https://www.manhattanreview.com/sat-practice-questions/}. Problems were presented in a random order during the main phase and in a fixed order during the final
phase.

\paragraph{Pretest Problem:}

One practice problem was presented to all participants. Neither the AI chat assistant nor the reference hints panel was
available during this phase.

\begin{graybox}
\textbf{Text 1.} Photo essays are increasingly popular in news outlets, but they
should not be classified as journalism. By definition, journalism conveys information
through language alone; photo essays communicate through images and use language only
sparingly, in captions and short blurbs. Readers experience a photo essay as a
sequence of pictures rather than as written reporting, which makes the form closer to
gallery art than to journalism. In this view, the core of journalism is extended
prose that explains events and evidence, so photo essays fall outside that category.

\medskip
\textbf{Text 2.} Photo essays present their reporting through both language and
images. Without captions, sequencing notes, and short text blocks, readers would not
know who is pictured, what happened, or why it matters. The account emerges from the
interaction of text and image. Moreover, acclaimed photo essays in major newspapers
include writing that meets the standards of careful reporting and clear style.
Therefore, photo essays qualify as journalistic work.

\medskip
\textit{Question:} Based on the texts, how would the author of Text 2 most likely
respond to the overall argument presented in Text 1?

\begin{enumerate}[label=\Alph*)]
  \item By arguing that journalism is not confined to prose alone and that photo
    essays, which use captions and short text to convey essential facts, satisfy
    accepted journalistic standards.
  \item By acknowledging that Text 1 correctly identifies a universal weakness in
    photo essays, namely that they lack sufficient language to report facts and
    therefore fail to meet journalistic standards.
  \item By suggesting that some photo essays are harder to follow than Text 1 admits,
    which complicates public reception, but leaves them better treated as a distinct
    visual form, not as journalism.
  \item By agreeing that photo essays lack rigorous reporting and writing comparable
    to standard articles and should therefore be regarded primarily as visual art
    rather than as journalism in professional contexts.
\end{enumerate}
\end{graybox}

\paragraph{Main Problems (5 problems, randomised order):}

The AI chat assistant was available to participants in the AI-assisted condition; the
reference hints panel was available to participants in the control condition.

\textbf{Problem 1.}
\begin{graybox}

\textbf{Text 1.} Esports draw huge audiences online and in arenas, but they should
not be classified as sports. By definition, sport is a physical contest in which
trained bodies display strength, endurance, or speed. Video game tournaments rely on
software, strategy, and quick thinking, with only limited bodily exertion. They
borrow the language of athletics, yet center on screen events rather than on-field
performance. In this sense, esports resemble chess or music performance more than
track or soccer, and they therefore fall outside the category of sport.

\medskip
\textbf{Text 2.} Supporters counter that esports present competitive events through
both coordinated physical control and rapid decision making. Without precise hand and
arm movements, players cannot execute tactics at all. Standardized rules, officiating,
and time limits govern matches, and teams follow rigorous practice schedules that
sport scientists now study in labs. Moreover, many accepted sports, such as archery
and shooting, prize accuracy and composure over heavy exertion. On these grounds,
esports satisfy the criteria for a sporting contest.

\medskip
\textit{Question:} Based on the texts, how would the author of Text 2 most likely
respond to the overall argument presented in Text 1?

\begin{enumerate}[label=\Alph*)]
  \item By asserting that criteria beyond raw exertion help define sport more than
    Text 1 admits, pointing to precision events and to esports' formal rules,
    training schedules, and objective competition.
  \item By acknowledging that Text 1 has exposed a universal flaw in esports, namely
    the lack of uniform rules, certified officials, and stable leagues that would
    justify any claim to sport.
  \item By suggesting that some esports possess rule systems far more complex than
    the author of Text 1 concedes, which means their difficulty warrants separate
    recognition rather than classification as sport.
  \item By agreeing that esports lack the rigorous physical training seen in
    established athletics and therefore should be considered entertainment products
    instead of sports within educational or professional contexts.
\end{enumerate}
\end{graybox}

\textbf{Problem 2.}
\begin{graybox}

\textbf{Text 1.} Anthropologist Maya Laird and colleagues measured carbon and
nitrogen isotope ratios in bone collagen from burials in two river valleys. They
reported $\delta^{13}$C and $\delta^{15}$N values consistent with heavy reliance on
maize by about 500 CE. Laird's team argues that these signatures support a
regionwide shift to intensive maize farming and storage, broadly similar to later
historical accounts. The authors contend that isotopes provide the clearest line of
evidence available for diet and land use at this time, implying that communities had
already adopted field systems that could sustain large settlements.

\medskip
\textbf{Text 2.} Archaeologists David Corbett and Lina Cho maintain that bulk
collagen isotopes are often overinterpreted as markers of maize agriculture. Similar
values can arise from marine fish consumption, arid-zone plants, or trophic-level
changes in freshwater settings. They note that Laird's study did not incorporate
compound-specific amino-acid analyses, enamel carbonate data, or plant microfossils
from tools and dental calculus, which could separate these possibilities. Corbett and
Cho conclude that without such complementary evidence, claims about an early,
regionwide maize regime likely rest on an incomplete reading of the available
signals.

\medskip
\textit{Question:} Based on the texts, how would Corbett and Cho (Text 2) most
likely characterize the conclusion presented in Text 1?

\begin{enumerate}[label=\Alph*)]
  \item As unexpected, because many scholars had believed maize was absent locally at
    that time, so the isotope-based claim sharply overturns prevailing views.
  \item As premature, because researchers in this region have only recently begun
    using isotope methods, leaving too little baseline information for firm dietary
    inferences.
  \item As questionable, because it relies on isotope signals that could reflect
    several alternative sources, and it omits complementary tests needed to
    distinguish among them.
  \item As puzzling, because it is rare to recover measurable collagen isotope values
    from burials of that age, making any dietary reconstruction unusually uncertain.
\end{enumerate}
\end{graybox}

\textbf{Problem 3.}
\begin{graybox}

\textbf{Text 1.} Astronomers have long wondered how tens of thousands of asteroids
can persist in the main belt between Mars and Jupiter while orbiting the same star
and drawing on the same material. According to conventional wisdom, gravity-driven
accretion should let a few bodies sweep up the rest. So why do so many remain as
separate objects after billions of years? Despite many proposed models, researchers
still have not reached a fully satisfying explanation.

\medskip
\textbf{Text 2.} Dynamical astronomer Leena Varma and colleagues connect the belt's
diversity to orbital stirring. Because these rocky bodies are small and widely
spaced, they rarely meet at gentle speeds. Jupiter's resonances continually jostle
their paths, turning potential mergers into shattering impacts. In such conditions,
direct growth by sticking probably happens far less than previously believed, which
reduces the kind of head-to-head competition that the older picture assumed.

\medskip
\textit{Question:} Based on the texts, how would Varma and colleagues (Text 2) most
likely respond to the ``conventional wisdom'' discussed in Text 1?

\begin{enumerate}[label=\Alph*)]
  \item By arguing that it rests on a misconception about how often asteroids can
    merge with one another in the stirred belt environment.
  \item By asserting that it ignores the continual delivery of fresh fragments from
    comets that prevents competition among belt asteroids.
  \item By suggesting that their own findings help clarify how asteroids are able to
    compete directly with Jupiter's powerful gravitational field.
  \item By recommending that more astronomers examine how merger rates among asteroids
    are increased when their typical encounter speeds become higher.
\end{enumerate}
\end{graybox}

\textbf{Problem 4.}
\begin{graybox}

\textbf{Text 1.} For decades, transportation policy followed a simple rule. If a
highway corridor was congested, add lanes and traffic would ease. Early postwar
projects seemed to confirm this view, with new pavement delivering smoother flow and
shorter trips. The apparent success fostered confidence that congestion falls in
predictable stages as capacity increases, a belief that guided many large road
programs. Yet chronic delays persisted in growing metro areas, prompting recurring
debates over whether the old recipe still captures how drivers and networks actually
behave.

\medskip
\textbf{Text 2.} In a recent synthesis, transportation researchers Maria Chen and
Raul Ortega argue that travel adapts quickly to fresh capacity. Using sensor data,
toll records, and long panel studies, they report that added lanes often attract new
trips, rerouted drivers, and longer distances within a few years. Some corridors show
brief relief followed by a rebound to familiar delays. The authors conclude that the
relationship between capacity and congestion is not one direction, and that under
common conditions the extra space restores queues rather than eliminating them.

\medskip
\textit{Question:} Based on the texts, how would Chen and Ortega (Text 2) most
likely respond to the ``conventional wisdom'' discussed in Text 1?

\begin{enumerate}[label=\Alph*)]
  \item By conceding that adding lanes sometimes helps, while asserting that transit
    and pricing are the more important long-term remedies for persistent urban
    congestion.
  \item By disputing the claim that congestion reliably declines in a step-by-step
    manner as capacity expands, given evidence that new lanes quickly fill with
    traffic.
  \item By acknowledging that induced demand likely was not a significant factor
    before smartphones and ride-hailing, which the texts suggest created large numbers
    of additional trips.
  \item By challenging the assumption that early downtown bottlenecks were the
    original causes of urban delay, rather than later suburban interchanges and
    peripheral ring roads.
\end{enumerate}
\end{graybox}

\textbf{Problem 5.}
\begin{graybox}

\textbf{Text 1.} Urban ecologists have long wondered how both crows and ravens can
thrive in the same cities while feeding on many of the same resources. According to
conventional wisdom, one species should eventually displace the other after
outcompeting it for food. Yet both birds continue to expand across large metropolitan
regions. Field surveys and casual observations have produced many partial
explanations, but none has fully solved the puzzle of coexistence. If crows and
ravens truly overlap in diet and habitat, why does one not prevail? Researchers still
lack a satisfactory account that explains how closely related scavengers persist side
by side in crowded urban landscapes.

\medskip
\textbf{Text 2.} Ornithologist Priya Raman and colleagues connect this coexistence
to fine-scale separation in time and space. Using GPS tags, camera traps, and stable
isotope analyses, they found that crows concentrate around schoolyards and parks
during daylight, while ravens favor early morning loading docks and taller rooftops.
The birds also differ in flight height and perch choice, which limits direct
encounters at food sources. Raman's team concludes that apparent overlap masks
reduced interaction. In cities, microhabitat and daily schedule differences likely
make head-to-head competition between crows and ravens much less common than
previously assumed.

\medskip
\textit{Question:} Based on the texts, how would Raman and colleagues (Text 2) most
likely respond to the ``conventional wisdom'' discussed in Text 1?

\begin{enumerate}[label=\Alph*)]
  \item By arguing that it rests on a misconception about how often crows and ravens
    actually compete directly for the same urban food sources.
  \item By asserting that it overlooks the steady influx of human food waste that
    prevents competition from emerging between the two corvid species.
  \item By suggesting that their findings mainly clarify how crows and ravens are
    able to compete successfully with larger city-dwelling raptors.
  \item By recommending that more researchers study how competition between crows and
    ravens increases when artificial lighting expands nighttime activity.
\end{enumerate}
\end{graybox}

\paragraph{Final Problems (Test Phase, 3 problems, fixed order):}

Neither the AI chat assistant nor the reference hints panel was available during this
phase for either condition.

\textbf{Problem 1.}
\begin{graybox}

\textbf{Text 1.} When cities convert mixed-traffic lanes into protected bus lanes,
officials often claim that commuters will benefit through faster, more reliable trips.
Transportation analyst Renee Solis tested this notion on a busy Midtown corridor by
modeling schedules, signal timing, and turning movements. Her simulations, calibrated
with probe-vehicle data, indicated that average door-to-door travel times for drivers
and for many bus riders would rise in the first year after the conversion. The modeled
delays reflected new turn restrictions, longer pedestrian phases, and spillback at
intersections that blocked side streets. Solis concluded that, contrary to the stated
goal, creating a bus-only facility on that corridor would likely make most commuters
worse off.

\medskip
\textbf{Text 2.} Urban economists Malik Ortiz and Sora Kim argue that evaluations of
street reallocations fixate on short windows and single-corridor metrics. Drawing on
multiyear data from several European and Latin American cities, they show that
protected bus lanes can trigger network adaptations: riders shift routes, agencies
retime signals, and drivers change departure times or modes. After these adjustments,
average person-throughput and trip reliability tend to improve citywide. Ortiz and
Kim contend that efficiency gains materialize over several years, so near-term
corridor slowdowns should not be taken as the final word on commuter welfare.

\medskip
\textit{Question:} Based on the texts, how would Ortiz and Kim (Text 2) most likely
respond to Solis's findings (Text 1)?

\begin{enumerate}[label=\Alph*)]
  \item They would recommend that Solis compare the near-term effect on Midtown
    travel times with results from a different corridor in another city.
  \item They would argue that, over the long term, operating costs for the bus system
    will also rise, offsetting any efficiency gains from the lane conversion.
  \item They would encourage Solis to investigate whether the modeled travel-time
    increases persist after network adaptations over an extended period.
  \item They would claim that protected bus lanes have fundamentally different effects
    in North American cities than in the cities used in their studies.
\end{enumerate}
\end{graybox}

\textbf{Problem 2.}
\begin{graybox}

\textbf{Text 1.} Public libraries that drop late fees often promise broader access,
but the policy can carry costs. Analyst Marcia Levin modeled a large city system's
first-year shift to fine-free lending. Calibrating with transaction logs and staff
time studies, she projected lower revenue, longer queues for popular titles, and a
small rise in very overdue items. Added customer service calls and reminder notices
absorbed hours once spent on programs. Levin concluded that, at least in the near
term, eliminating fines would leave patrons waiting longer while the system faces
tighter budgets and higher operating burdens. Her report urged caution before scaling
the policy beyond a pilot.

\medskip
\textbf{Text 2.} Library researchers Damon Ruiz and Alisha Porter argue that
evaluations of fine-free policies stop too soon. Using multiyear evidence from dozens
of systems, they find that card renewals rise, materials are returned at higher rates
after improved outreach, and circulation expands in neighborhoods with past debt
blocks. Processing costs fall when staff stop handling coin boxes and appeals, and
grants and small donations offset lost fees. After two to three budget cycles, per-loan
costs and holds stabilize. Ruiz and Porter contend that early disruptions are real but
temporary, and that long-run patron engagement and efficiency gains should guide
decisions.

\medskip
\textit{Question:} Based on the texts, how would Ruiz and Porter (Text 2) most
likely respond to Levin's findings (Text 1)?

\begin{enumerate}[label=\Alph*)]
  \item They would suggest comparing the short-term effects in Levin's city with
    outcomes from another city that kept fines, to control for differences in catalog
    size and local demand.
  \item They would argue that, over time, operating costs for a fine-free system
    increase further, which cancels out any gains in access that patrons might
    experience after the policy change.
  \item They would encourage Levin to test whether the projected delays and budget
    pressures persist after two or more years of outreach, process changes, and
    broader patron re-engagement.
  \item They would claim that fine-free policies have fundamentally different effects
    in large urban systems than in the smaller systems that appear in their multiyear
    datasets.
\end{enumerate}
\end{graybox}

\textbf{Problem 3.}
\begin{graybox}

\textbf{Text 1.} City bridges eventually crack from stress and weather. But
researchers studying ancient Roman harbor ruins found that their concrete seems to
``heal'' small fractures when seawater seeps in. Recent lab work reports a similar
effect in modern trial mixes that include tiny lime-rich clasts and pozzolanic ash,
with microcracks closing over weeks in brine. Engineers say this parallel points to
an accessible path to longer-lasting highway bridges, potentially reducing maintenance
closures and costly emergency repairs.

\medskip
\textbf{Text 2.} When materials scientist Lila Moreno and her team described
healing-like behavior in their modern mixes, the claim drew interest. Yet, as they
emphasized, the chemistry appears to depend on slow marine interactions not typical
of bridge decks, and the role of the clasts is not fully established. Moreno has
compared the clasts to ``seeds'' that may release binding material under certain
conditions, but how this would work in everyday bridge environments remains uncertain.

\medskip
\textit{Question:} Based on the texts, what would the author of Text 2 most likely
say about Text 1's characterization of the modern-concrete finding?

\begin{enumerate}[label=\Alph*)]
  \item It is reasonable given the team has confirmed the same mechanism in both
    marine settings and highway bridges.
  \item It is overly optimistic because the mechanism and its relevance to typical
    bridge conditions remain uncertain.
  \item It is unexpected because the initial report received little enthusiasm from
    materials scientists.
  \item It is unfairly dismissive because Text 1 ignores successful large-scale
    highway trials already completed.
\end{enumerate}
\end{graybox}

\subsection{Reference Materials Available to Participants}\label{App:reading_sidebar}

\paragraph{Control condition: Reference hints panel.}

Participants in the control condition had access to a fixed sidebar
panel titled \textit{``Reading Comprehension Tips''} throughout the main experimental
phase. The panel contained the following tips:

\begin{graybox}
\begin{itemize}
  \item \textbf{Read the Question First:} Understand what you are looking for (main
    purpose, function of a sentence, main idea, etc.).
  \item \textbf{Context is Key:} Look at the surrounding sentences to understand the
    role of specific parts of the text.
  \item \textbf{Eliminate Wrong Answers:} Often, choices are partially correct but
    contain one detail that makes them wrong.
\end{itemize}
\end{graybox}

\paragraph{AI-assisted condition: Chat assistant greeting.}

Participants in the AI-assisted condition had access to a chat interface (described
to participants as ChatGPT) throughout the main experimental phase. The assistant
opened each problem with the following greeting:

\begin{graybox}
\textit{``Hi! I'm here to help you with this reading comprehension question. I can
see the text and the question. Feel free to ask anything!''}
\end{graybox}

The assistant had access to the passage text, the question, and all four answer
options for the currently displayed problem, and could respond to any free-text
queries from the participant.

\subsection{Attrition Analysis}\label{App:attrition}

\begin{figure}[t!]
\centering

\begin{subfigure}[t]{0.49\linewidth}
    \centering
    \includegraphics[width=\linewidth]{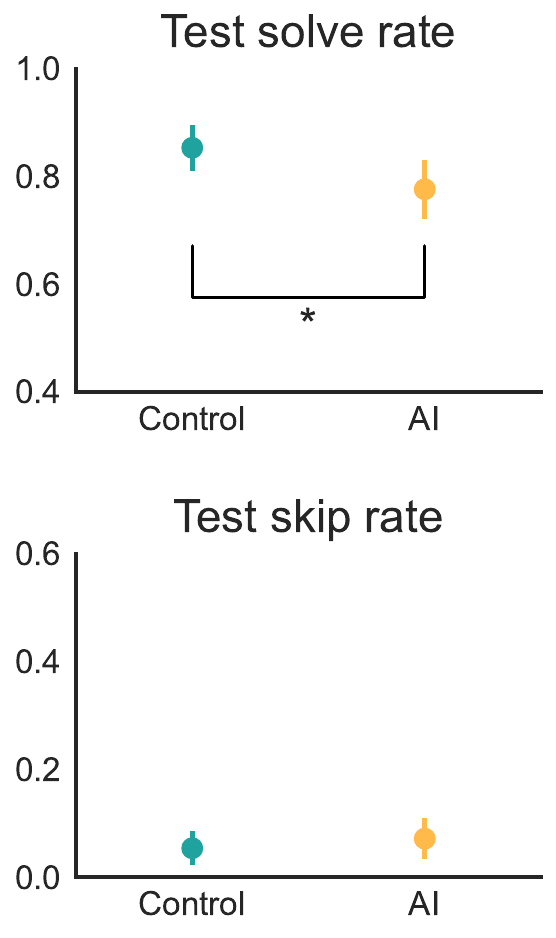}
    \caption{}
    \label{fig:exp2_attrition}
\end{subfigure}
\hfill
\begin{subfigure}[t]{0.49\linewidth}
    \centering
    \includegraphics[width=\linewidth]{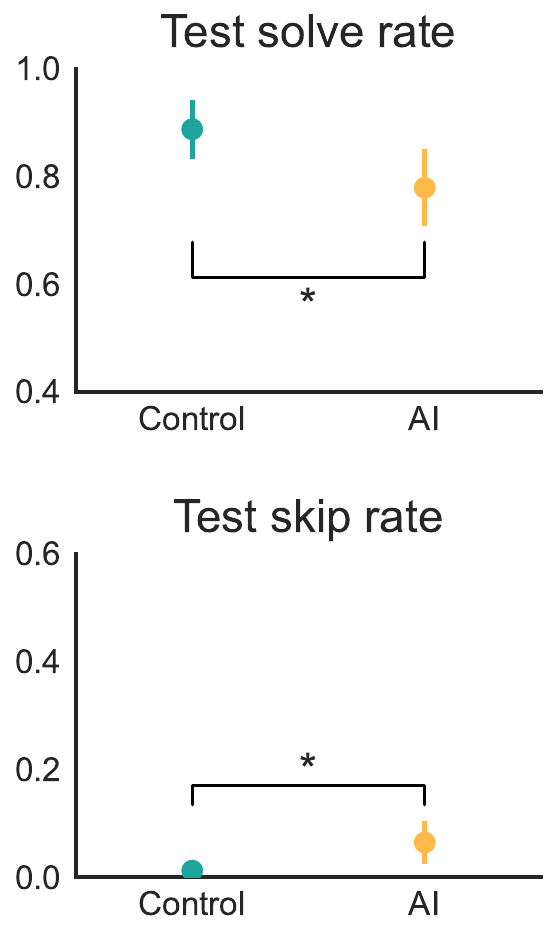}
    \caption{}
    \label{fig:exp3_attrition}
\end{subfigure}

\caption{
\textbf{Performance declines remain after controlling for potential attrition effects.}
(a) Experiment 2: The performance decline holds when restricting the analysis to participants who solved all three pretest questions.
(b) Experiment 3: The performance and persistence declines hold when excluding the lowest-scoring AI participants to exactly match the control sample size.
}
\label{fig:attrition_analysis}
\end{figure}

To address a potential difference in attrition rates in Experiment 2 and Experiment 3, we conduct additional analysis. For experiment 2, we filter specifically for highly motivated/skilled participants who solved all 3 pre-test questions correctly. Indeed, this filter excluded more AI participants (207/349 excluded) than control participants (173/318 excluded), leaving a final split of 143 AI to 140 control participants. We find that the gap in test performance remains (Fig~\ref{fig:exp2_attrition}), with a mean solve rate of $0.85$ for control participants and $0.78$ for AI particiants ($P=0.03$). Similarly to the main results for Experiment 2 (section 3.2), the skip rate is higher for AI participants (mean $0.07$) than for control participants (mean $0.05$), though the difference is not significant ($P=0.47$).

In Experiment 3, we only have 1 pretest question, which we used as our default exclusion criteria. After excluding participants who did not solve the pretest question, there were 85 AI participants and 83 control participants. To address your concern, we then excluded the two participants in the AI condition that had the lowest test solve rate leaving 83 participants in both the AI and control condition. We did this as a conservative analysis and to ensure that the sample sizes for both conditions are matched. 

We find that the effect holds in this analysis (Fig~\ref{fig:exp3_attrition}): Participants in the AI condition have a lower solve rate (mean $0.78$) compared to participants in the control condition (mean $0.89$, $P=0.018$). Additionally, participants in the AI condition have a higher skip rate (mean $0.06$) compared to participants in the control condition (mean $0.01$, $P=0.019$). 

\begin{thebibliography}{78}
\providecommand{\natexlab}[1]{#1}
\providecommand{\url}[1]{\texttt{#1}}
\expandafter\ifx\csname urlstyle\endcsname\relax
  \providecommand{\doi}[1]{doi: #1}\else
  \providecommand{\doi}{doi: \begingroup \urlstyle{rm}\Url}\fi

\bibitem[Andersson \& Bergman(2011)Andersson and Bergman]{andersson2011role}
H{\aa}kan Andersson and Lars~R Bergman.
\newblock The role of task persistence in young adolescence for successful educational and occupational attainment in middle adulthood.
\newblock \emph{Developmental psychology}, 47\penalty0 (4):\penalty0 950, 2011.

\bibitem[Balcazar \& Keys(2014)Balcazar and Keys]{balcazar2014goals}
Fabricio~E Balcazar and Christopher~B Keys.
\newblock Goals in mentoring relationships.
\newblock \emph{Handbook of Youth Mentoring}, 2, 2014.

\bibitem[Barcaui(2025)]{barcaui2025chatgpt}
Andr{\'e} Barcaui.
\newblock Chatgpt as a cognitive crutch: Evidence from a randomized controlled trial on knowledge retention.
\newblock \emph{Social Sciences \& Humanities Open}, 12:\penalty0 102287, 2025.

\bibitem[Bastani et~al.(2025)Bastani, Bastani, Sungu, Ge, Kabakc{\i}, and Mariman]{bastani2025generative}
Hamsa Bastani, Osbert Bastani, Alp Sungu, Haosen Ge, {\"O}zge Kabakc{\i}, and Rei Mariman.
\newblock Generative ai without guardrails can harm learning: Evidence from high school mathematics.
\newblock \emph{Proceedings of the National Academy of Sciences}, 122\penalty0 (26):\penalty0 e2422633122, 2025.

\bibitem[Bastug(2014)]{bastug2014structural}
Muhammet Bastug.
\newblock The structural relationship of reading attitude, reading comprehension and academic achievement.
\newblock \emph{International Journal of Social Sciences and Education}, 4\penalty0 (4):\penalty0 931--946, 2014.

\bibitem[Bengio et~al.(2024)Bengio, Hinton, Yao, Song, Abbeel, Darrell, Harari, Zhang, Xue, Shalev-Shwartz, et~al.]{bengio2024managing}
Yoshua Bengio, Geoffrey Hinton, Andrew Yao, Dawn Song, Pieter Abbeel, Trevor Darrell, Yuval~Noah Harari, Ya-Qin Zhang, Lan Xue, Shai Shalev-Shwartz, et~al.
\newblock Managing extreme {AI} risks amid rapid progress.
\newblock \emph{Science}, 384\penalty0 (6698):\penalty0 842--845, 2024.

\bibitem[Bjork et~al.(2011)Bjork, Bjork, et~al.]{bjork2011making}
Elizabeth~L Bjork, Robert~A Bjork, et~al.
\newblock Making things hard on yourself, but in a good way: Creating desirable difficulties to enhance learning.
\newblock \emph{Psychology and the real world: Essays illustrating fundamental contributions to society}, 2\penalty0 (59-68):\penalty0 56--64, 2011.

\bibitem[Braithwaite \& Siegler(2021)Braithwaite and Siegler]{braithwaite2021putting}
David~W Braithwaite and Robert~S Siegler.
\newblock Putting fractions together.
\newblock \emph{Journal of Educational Psychology}, 113\penalty0 (3):\penalty0 556, 2021.

\bibitem[Bratman(1992)]{bratman1992shared}
Michael~E Bratman.
\newblock Shared cooperative activity.
\newblock \emph{The philosophical review}, 101\penalty0 (2):\penalty0 327--341, 1992.

\bibitem[Brickman(1971)]{brickman1971hedonic}
Philip Brickman.
\newblock Hedonic relativism and planning the good society.
\newblock \emph{Adaptation level theory}, pp.\  287--301, 1971.

\bibitem[Brickman et~al.(1978)Brickman, Coates, and Janoff-Bulman]{brickman1978lottery}
Philip Brickman, Dan Coates, and Ronnie Janoff-Bulman.
\newblock Lottery winners and accident victims: Is happiness relative?
\newblock \emph{Journal of personality and social psychology}, 36\penalty0 (8):\penalty0 917, 1978.

\bibitem[Brynjolfsson et~al.(2025)Brynjolfsson, Li, and Raymond]{brynjolfsson2025generative}
Erik Brynjolfsson, Danielle Li, and Lindsey Raymond.
\newblock Generative {AI} at work.
\newblock \emph{The Quarterly Journal of Economics}, 140\penalty0 (2):\penalty0 889--942, 2025.

\bibitem[Bu{\c{c}}inca et~al.(2024)Bu{\c{c}}inca, Swaroop, Paluch, Murphy, and Gajos]{buccinca2024towards}
Zana Bu{\c{c}}inca, Siddharth Swaroop, Amanda~E Paluch, Susan~A Murphy, and Krzysztof~Z Gajos.
\newblock Towards optimizing human-centric objectives in {AI}-assisted decision-making with offline reinforcement learning.
\newblock \emph{arXiv preprint arXiv:2403.05911}, 2024.

\bibitem[Bucknall \& Dori-Hacohen(2022)Bucknall and Dori-Hacohen]{bucknall2022current}
Benjamin~S Bucknall and Shiri Dori-Hacohen.
\newblock Current and near-term {AI} as a potential existential risk factor.
\newblock In \emph{Proceedings of the 2022 AAAI/ACM Conference on AI, Ethics, and Society}, pp.\  119--129, 2022.

\bibitem[Budzy{\'n} et~al.(2025)Budzy{\'n}, Roma{\'n}czyk, Kitala, Ko{\l}odziej, Bugajski, Adami, Blom, Buszkiewicz, Halvorsen, Hassan, et~al.]{budzyn2025endoscopist}
Krzysztof Budzy{\'n}, Marcin Roma{\'n}czyk, Diana Kitala, Pawe{\l} Ko{\l}odziej, Marek Bugajski, Hans~O Adami, Johannes Blom, Marek Buszkiewicz, Natalie Halvorsen, Cesare Hassan, et~al.
\newblock Endoscopist deskilling risk after exposure to artificial intelligence in colonoscopy: a multicentre, observational study.
\newblock \emph{The Lancet Gastroenterology \& Hepatology}, 10\penalty0 (10):\penalty0 896--903, 2025.

\bibitem[Carlsmith(2022)]{carlsmith2022power}
Joseph Carlsmith.
\newblock Is power-seeking {AI} an existential risk?
\newblock \emph{arXiv preprint arXiv:2206.13353}, 2022.

\bibitem[Cheng et~al.(2025{\natexlab{a}})Cheng, Lee, Khadpe, Yu, Han, and Jurafsky]{cheng2025sycophantic}
Myra Cheng, Cinoo Lee, Pranav Khadpe, Sunny Yu, Dyllan Han, and Dan Jurafsky.
\newblock Sycophantic {AI} decreases prosocial intentions and promotes dependence.
\newblock \emph{arXiv preprint arXiv:2510.01395}, 2025{\natexlab{a}}.

\bibitem[Cheng et~al.(2025{\natexlab{b}})Cheng, Yu, Lee, Khadpe, Ibrahim, and Jurafsky]{cheng2025social}
Myra Cheng, Sunny Yu, Cinoo Lee, Pranav Khadpe, Lujain Ibrahim, and Dan Jurafsky.
\newblock Social sycophancy: A broader understanding of llm sycophancy.
\newblock \emph{arXiv preprint arXiv:2505.13995}, 2025{\natexlab{b}}.

\bibitem[Cheng et~al.(2026)Cheng, Hawkins, and Jurafsky]{cheng2026accommodation}
Myra Cheng, Robert~D Hawkins, and Dan Jurafsky.
\newblock Accommodation and epistemic vigilance: A pragmatic account of why llms fail to challenge harmful beliefs.
\newblock \emph{arXiv preprint arXiv:2601.04435}, 2026.

\bibitem[Christian(2025)]{christian2025computational}
Brian Christian.
\newblock Computational frameworks for human care.
\newblock \emph{D{\ae}dalus}, 154\penalty0 (1):\penalty0 183--197, 2025.

\bibitem[Christian et~al.(2026)Christian, Thompson, Yang, Adam, Kirk, Summerfield, and Dumbalska]{christian2026reward}
Brian Christian, Jessica~AF Thompson, Elle~Michelle Yang, Vincent Adam, Hannah~Rose Kirk, Christopher Summerfield, and Tsvetomira Dumbalska.
\newblock Reward models inherit value biases from pretraining.
\newblock \emph{arXiv preprint arXiv:2601.20838}, 2026.

\bibitem[Collins et~al.(2024)Collins, Sucholutsky, Bhatt, Chandra, Wong, Lee, Zhang, Zhi-Xuan, Ho, Mansinghka, et~al.]{collins2024building}
Katherine~M Collins, Ilia Sucholutsky, Umang Bhatt, Kartik Chandra, Lionel Wong, Mina Lee, Cedegao~E Zhang, Tan Zhi-Xuan, Mark Ho, Vikash Mansinghka, et~al.
\newblock Building machines that learn and think with people.
\newblock \emph{Nature Human Behaviour}, 8\penalty0 (10):\penalty0 1851--1863, 2024.

\bibitem[Doshi \& Hauser(2024)Doshi and Hauser]{doshi2024generative}
Anil~R Doshi and Oliver~P Hauser.
\newblock Generative {AI} enhances individual creativity but reduces the collective diversity of novel content.
\newblock \emph{Science advances}, 10\penalty0 (28):\penalty0 eadn5290, 2024.

\bibitem[Dubey et~al.(2021)Dubey, Ho, Mehta, and Griffiths]{dubey2021aha}
Rachit Dubey, Mark Ho, Hermish Mehta, and Tom Griffiths.
\newblock Aha! moments correspond to metacognitive prediction errors.
\newblock \emph{Available at SSRN 5475080}, 2021.

\bibitem[Duckworth et~al.(2007)Duckworth, Peterson, Matthews, and Kelly]{duckworth2007grit}
Angela~L Duckworth, Christopher Peterson, Michael~D Matthews, and Dennis~R Kelly.
\newblock Grit: perseverance and passion for long-term goals.
\newblock \emph{Journal of Personality and Social Psychology}, 92\penalty0 (6):\penalty0 1087, 2007.

\bibitem[Elizondo et~al.(2024)Elizondo, Valenzuela, Pestana, and Codina]{elizondo2024self}
Karla Elizondo, Rafael Valenzuela, Jos{\'e}~V Pestana, and Nuria Codina.
\newblock Self-regulation and procrastination in college students: A tale of motivation, strategy, and perseverance.
\newblock \emph{Psychology in the Schools}, 61\penalty0 (3):\penalty0 887--902, 2024.

\bibitem[Fleming \& Daw(2017)Fleming and Daw]{fleming2017self}
Stephen~M Fleming and Nathaniel~D Daw.
\newblock Self-evaluation of decision-making: A general bayesian framework for metacognitive computation.
\newblock \emph{Psychological Review}, 124\penalty0 (1):\penalty0 91, 2017.

\bibitem[Frederick \& Loewenstein(1999)Frederick and Loewenstein]{frederick199916}
Shane Frederick and George Loewenstein.
\newblock Hedonic adaptation.
\newblock In Daniel Kahneman, Edward Diener, and Norbert Schwarz (eds.), \emph{Well-Being: The Foundations of Hedonic Psychology}, pp.\  302--329. 1999.

\bibitem[Gerlich(2025)]{gerlich2025ai}
Michael Gerlich.
\newblock {AI} tools in society: Impacts on cognitive offloading and the future of critical thinking.
\newblock \emph{Societies}, 15\penalty0 (1):\penalty0 6, 2025.

\bibitem[Gilbert et~al.(2020)Gilbert, Bird, Carpenter, Fleming, Sachdeva, and Tsai]{gilbert2020optimal}
Sam~J Gilbert, Arabella Bird, Jason~M Carpenter, Stephen~M Fleming, Chhavi Sachdeva, and Pei-Chun Tsai.
\newblock Optimal use of reminders: Metacognition, effort, and cognitive offloading.
\newblock \emph{Journal of Experimental Psychology: General}, 149\penalty0 (3):\penalty0 501, 2020.

\bibitem[Goldin-Meadow et~al.(2001)Goldin-Meadow, Nusbaum, Kelly, and Wagner]{goldin2001explaining}
Susan Goldin-Meadow, Howard Nusbaum, Spencer~D Kelly, and Susan Wagner.
\newblock Explaining math: Gesturing lightens the load.
\newblock \emph{Psychological Science}, 12\penalty0 (6):\penalty0 516--522, 2001.

\bibitem[Grinschgl et~al.(2021)Grinschgl, Papenmeier, and Meyerhoff]{grinschgl2021consequences}
Sandra Grinschgl, Frank Papenmeier, and Hauke~S Meyerhoff.
\newblock Consequences of cognitive offloading: Boosting performance but diminishing memory.
\newblock \emph{Quarterly Journal of Experimental Psychology}, 74\penalty0 (9):\penalty0 1477--1496, 2021.

\bibitem[Grosz \& Kraus(1996)Grosz and Kraus]{grosz1996collaborative}
Barbara~J Grosz and Sarit Kraus.
\newblock Collaborative plans for complex group action.
\newblock \emph{Artificial Intelligence}, 86\penalty0 (2):\penalty0 269--357, 1996.

\bibitem[Guingrich et~al.(2026)Guingrich, Mehta, and Bhatt]{guingrich2026belief}
Rose~E Guingrich, Dvija Mehta, and Umang Bhatt.
\newblock Belief offloading in human-{AI} interaction.
\newblock \emph{arXiv preprint arXiv:2602.08754}, 2026.

\bibitem[Guiso et~al.(2016)Guiso, Sapienza, and Zingales]{guiso2016long}
Luigi Guiso, Paola Sapienza, and Luigi Zingales.
\newblock Long-term persistence.
\newblock \emph{Journal of the European Economic Association}, 14\penalty0 (6):\penalty0 1401--1436, 2016.

\bibitem[Ibrahim et~al.(2025)Ibrahim, Collins, Kim, Reuel, Lamparth, Feng, Ahmad, Soni, Kattan, Stein, et~al.]{ibrahim2025measuring}
Lujain Ibrahim, Katherine~M Collins, Sunnie~SY Kim, Anka Reuel, Max Lamparth, Kevin Feng, Lama Ahmad, Prajna Soni, Alia~El Kattan, Merlin Stein, et~al.
\newblock Measuring and mitigating overreliance is necessary for building human-compatible {AI}.
\newblock \emph{arXiv preprint arXiv:2509.08010}, 2025.

\bibitem[Jakesch et~al.(2023)Jakesch, Bhat, Buschek, Zalmanson, and Naaman]{jakesch2023co}
Maurice Jakesch, Advait Bhat, Daniel Buschek, Lior Zalmanson, and Mor Naaman.
\newblock Co-writing with opinionated language models affects users' views.
\newblock In \emph{Proceedings of the 2023 CHI Conference on Human Factors in Computing Systems}, pp.\  1--15, 2023.

\bibitem[Kapur(2014)]{kapur2014productive}
Manu Kapur.
\newblock Productive failure in learning math.
\newblock \emph{Cognitive Science}, 38\penalty0 (5):\penalty0 1008--1022, 2014.

\bibitem[Kasirzadeh(2025)]{kasirzadeh2025two}
Atoosa Kasirzadeh.
\newblock Two types of {AI} existential risk: decisive and accumulative: A. kasirzadeh.
\newblock \emph{Philosophical Studies}, 182\penalty0 (7):\penalty0 1975--2003, 2025.

\bibitem[Ke et~al.(2026)Ke, Jin, Ong, Thirunavukarasu, Car, Cheung, Tham, Ting, Ong, Compton, et~al.]{ke2026ai}
Yuhe Ke, Liyuan Jin, Jasmine Chiat~Ling Ong, Arun~J Thirunavukarasu, Josip Car, Carol~Y Cheung, Yih~Chung Tham, Daniel Shu~Wei Ting, Marcus Eng~Hock Ong, Scott Compton, et~al.
\newblock Ai-induced never-skilling in medical education.
\newblock \emph{Nature medicine}, pp.\  1--10, 2026.

\bibitem[Kim et~al.(2025)Kim, Vaughan, Liao, Lombrozo, and Russakovsky]{kim2025fostering}
Sunnie~SY Kim, Jennifer~Wortman Vaughan, Q~Vera Liao, Tania Lombrozo, and Olga Russakovsky.
\newblock Fostering appropriate reliance on large language models: The role of explanations, sources, and inconsistencies.
\newblock In \emph{Proceedings of the 2025 CHI Conference on Human Factors in Computing Systems}, pp.\  1--19, 2025.

\bibitem[Kintsch(1998)]{kintsch1998comprehension}
Walter Kintsch.
\newblock \emph{Comprehension: A paradigm for cognition}.
\newblock Cambridge university press, 1998.

\bibitem[Kirk et~al.(2025{\natexlab{a}})Kirk, Davidson, Saunders, Luettgau, Vidgen, Hale, and Summerfield]{kirk2025neural}
Hannah~Rose Kirk, Henry Davidson, Ed~Saunders, Lennart Luettgau, Bertie Vidgen, Scott~A Hale, and Christopher Summerfield.
\newblock Neural steering vectors reveal dose and exposure-dependent impacts of human-ai relationships.
\newblock \emph{arXiv preprint arXiv:2512.01991}, 2025{\natexlab{a}}.

\bibitem[Kirk et~al.(2025{\natexlab{b}})Kirk, Gabriel, Summerfield, Vidgen, and Hale]{kirk2025human}
Hannah~Rose Kirk, Iason Gabriel, Chris Summerfield, Bertie Vidgen, and Scott~A Hale.
\newblock Why human--ai relationships need socioaffective alignment.
\newblock \emph{Humanities and Social Sciences Communications}, 12\penalty0 (1):\penalty0 1--9, 2025{\natexlab{b}}.

\bibitem[Kleiman-Weiner(2024)]{kleiman2024computational}
Max Kleiman-Weiner.
\newblock Computational principles of caregiving.
\newblock In \emph{Proceedings of the Annual Meeting of the Cognitive Science Society}, volume~46, 2024.

\bibitem[Koedinger \& Aleven(2007)Koedinger and Aleven]{koedinger2007exploring}
Kenneth~R Koedinger and Vincent Aleven.
\newblock Exploring the assistance dilemma in experiments with cognitive tutors.
\newblock \emph{Educational Psychology Review}, 19\penalty0 (3):\penalty0 239--264, 2007.

\bibitem[Kosmyna et~al.(2025)Kosmyna, Hauptmann, Yuan, Situ, Liao, Beresnitzky, Braunstein, and Maes]{kosmyna2025your}
Nataliya Kosmyna, Eugene Hauptmann, Ye~Tong Yuan, Jessica Situ, Xian-Hao Liao, Ashly~Vivian Beresnitzky, Iris Braunstein, and Pattie Maes.
\newblock Your brain on chatgpt: Accumulation of cognitive debt when using an {AI} assistant for essay writing task.
\newblock \emph{arXiv preprint arXiv:2506.08872}, 4, 2025.

\bibitem[Kulveit et~al.(2025)Kulveit, Douglas, Ammann, Turan, Krueger, and Duvenaud]{kulveit2025gradual}
Jan Kulveit, Raymond Douglas, Nora Ammann, Deger Turan, David Krueger, and David Duvenaud.
\newblock Gradual disempowerment: Systemic existential risks from incremental {AI} development.
\newblock \emph{arXiv preprint arXiv:2501.16946}, 2025.

\bibitem[Lee et~al.(2025)Lee, Sarkar, Tankelevitch, Drosos, Rintel, Banks, and Wilson]{lee2025impact}
Hao-Ping Lee, Advait Sarkar, Lev Tankelevitch, Ian Drosos, Sean Rintel, Richard Banks, and Nicholas Wilson.
\newblock The impact of generative {AI} on critical thinking: Self-reported reductions in cognitive effort and confidence effects from a survey of knowledge workers.
\newblock In \emph{Proceedings of the 2025 CHI Conference on Human Factors in Computing Systems}, pp.\  1--22, 2025.

\bibitem[Liu et~al.(2025)Liu, Snell, Griffiths, and Dubey]{liu2025binary}
Grace Liu, Jake~C Snell, Thomas~L Griffiths, and Rachit Dubey.
\newblock Binary climate data visuals amplify perceived impact of climate change.
\newblock \emph{Nature Human Behaviour}, 9\penalty0 (7):\penalty0 1355--1364, 2025.

\bibitem[Maddux(2009)]{maddux200931}
James~E Maddux.
\newblock Self-efficacy: The power of believing you can.
\newblock In \emph{The Oxford Handbook of Positive Psychology}, pp.\  335--344. Oxford University Press, 2009.

\bibitem[Martin \& Schwartz(2005)Martin and Schwartz]{martin2005physically}
Taylor Martin and Daniel~L Schwartz.
\newblock Physically distributed learning: Adapting and reinterpreting physical environments in the development of fraction concepts.
\newblock \emph{Cognitive Science}, 29\penalty0 (4):\penalty0 587--625, 2005.

\bibitem[Mattessich \& Johnson(2018)Mattessich and Johnson]{mattessich2018collaboration}
Paul~W. Mattessich and Kirsten~M. Johnson.
\newblock \emph{Collaboration: What Makes It Work}.
\newblock Turner Publishing Company, 3rd edition, 2018.

\bibitem[Metcalfe(2009)]{metcalfe2009metacognitive}
Janet Metcalfe.
\newblock Metacognitive judgments and control of study.
\newblock \emph{Current Directions in Psychological Science}, 18\penalty0 (3):\penalty0 159--163, 2009.

\bibitem[Metcalfe \& Mischel(1999)Metcalfe and Mischel]{metcalfe1999hot}
Janet Metcalfe and Walter Mischel.
\newblock A hot/cool-system analysis of delay of gratification: dynamics of willpower.
\newblock \emph{Psychological Review}, 106\penalty0 (1):\penalty0 3, 1999.

\bibitem[Mooradian et~al.(2016)Mooradian, Matzler, Uzelac, and Bauer]{mooradian2016perspiration}
Todd Mooradian, Kurt Matzler, Borislav Uzelac, and Florian Bauer.
\newblock Perspiration and inspiration: Grit and innovativeness as antecedents of entrepreneurial success.
\newblock \emph{Journal of Economic Psychology}, 56:\penalty0 232--243, 2016.

\bibitem[Moore et~al.(2019)Moore, Obradovich, Lehner, and Baylis]{moore2019rapidly}
Frances~C Moore, Nick Obradovich, Flavio Lehner, and Patrick Baylis.
\newblock Rapidly declining remarkability of temperature anomalies may obscure public perception of climate change.
\newblock \emph{Proceedings of the National Academy of Sciences}, 116\penalty0 (11):\penalty0 4905--4910, 2019.

\bibitem[{OECD}(2026)]{oecd2026digital}
{OECD}.
\newblock \emph{{OECD} Digital Education Outlook 2026: Exploring Effective Uses of Generative {AI} in Education}.
\newblock OECD Publishing, Paris, 2026.
\newblock \doi{10.1787/062a7394-en}.
\newblock URL \url{https://doi.org/10.1787/062a7394-en}.

\bibitem[Padmakumar \& He(2023)Padmakumar and He]{padmakumar2023does}
Vishakh Padmakumar and He~He.
\newblock Does writing with language models reduce content diversity?
\newblock \emph{arXiv preprint arXiv:2309.05196}, 2023.

\bibitem[Pourhosein~Gilakjani \& Sabouri(2016)Pourhosein~Gilakjani and Sabouri]{pourhosein2016can}
Abbas Pourhosein~Gilakjani and Narjes~Banou Sabouri.
\newblock How can students improve their reading comprehension skill.
\newblock \emph{Journal of Studies in Education}, 6\penalty0 (2):\penalty0 229, 2016.

\bibitem[Rathje et~al.(2025)Rathje, Ye, Globig, Pillai, Oldemburgo~de Mello, and Van~Bavel]{rathje2025sycophantic}
Steve Rathje, Meryl Ye, Laura~K. Globig, Raunak~M. Pillai, Victoria Oldemburgo~de Mello, and Jay~J. Van~Bavel.
\newblock Sycophantic {AI} increases attitude extremity and overconfidence, 2025.
\newblock Preprint, PsyArXiv.

\bibitem[Richmond \& Taylor(2025)Richmond and Taylor]{richmond2025benefits}
Lauren~L Richmond and Ryan~G Taylor.
\newblock The benefits and potential costs of cognitive offloading for retrospective information.
\newblock \emph{Nature Reviews Psychology}, 4\penalty0 (5):\penalty0 312--321, 2025.

\bibitem[Risko \& Gilbert(2016)Risko and Gilbert]{risko2016cognitive}
Evan~F Risko and Sam~J Gilbert.
\newblock Cognitive offloading.
\newblock \emph{Trends in Cognitive Sciences}, 20\penalty0 (9):\penalty0 676--688, 2016.

\bibitem[Shapira et~al.(2026)Shapira, Benade, and Procaccia]{shapira2026rlhf}
Itai Shapira, Gerdus Benade, and Ariel~D Procaccia.
\newblock How {RLHF} amplifies sycophancy.
\newblock \emph{arXiv preprint arXiv:2602.01002}, 2026.

\bibitem[Sharma et~al.(2026)Sharma, McCain, Douglas, and Duvenaud]{sharma2026s}
Mrinank Sharma, Miles McCain, Raymond Douglas, and David Duvenaud.
\newblock Who's in charge? disempowerment patterns in real-world {LLM} usage.
\newblock \emph{arXiv preprint arXiv:2601.19062}, 2026.

\bibitem[Shen \& Tamkin(2026)Shen and Tamkin]{shen2026ai}
Judy~Hanwen Shen and Alex Tamkin.
\newblock How {AI} impacts skill formation.
\newblock \emph{arXiv preprint arXiv:2601.20245}, 2026.

\bibitem[Siegler \& Pyke(2013)Siegler and Pyke]{siegler2013developmental}
Robert~S Siegler and Aryn~A Pyke.
\newblock Developmental and individual differences in understanding of fractions.
\newblock \emph{Developmental Psychology}, 49\penalty0 (10):\penalty0 1994, 2013.

\bibitem[Soderstrom \& Bjork(2015)Soderstrom and Bjork]{soderstrom2015learning}
Nicholas~C Soderstrom and Robert~A Bjork.
\newblock Learning versus performance: An integrative review.
\newblock \emph{Perspectives on Psychological Science}, 10\penalty0 (2):\penalty0 176--199, 2015.

\bibitem[Sparrow et~al.(2011)Sparrow, Liu, and Wegner]{sparrow2011google}
Betsy Sparrow, Jenny Liu, and Daniel~M Wegner.
\newblock Google effects on memory: Cognitive consequences of having information at our fingertips.
\newblock \emph{Science}, 333\penalty0 (6043):\penalty0 776--778, 2011.

\bibitem[Steyvers \& Mayer(2025)Steyvers and Mayer]{steyvers2025not}
Mark Steyvers and Lukas Mayer.
\newblock When not to help: planning for lasting human-{AI} collaboration.
\newblock \emph{arXiv preprint arXiv:2508.01837}, 2025.

\bibitem[Sucholutsky et~al.(2025)Sucholutsky, Collins, Jacoby, Thompson, and Hawkins]{sucholutsky2025using}
Ilia Sucholutsky, Katherine~M Collins, Nori Jacoby, Bill~D Thompson, and Robert~D Hawkins.
\newblock Using llms to advance the cognitive science of collectives.
\newblock \emph{Nature Computational Science}, 5\penalty0 (9):\penalty0 704--707, 2025.

\bibitem[Sumers et~al.(2022)Sumers, Hawkins, Ho, Griffiths, and Hadfield-Menell]{sumers2022talk}
Theodore Sumers, Robert Hawkins, Mark~K Ho, Tom Griffiths, and Dylan Hadfield-Menell.
\newblock How to talk so {AI} will learn: Instructions, descriptions, and autonomy.
\newblock \emph{Advances in Neural Information Processing Systems}, 35:\penalty0 34762--34775, 2022.

\bibitem[Van~de Pol et~al.(2010)Van~de Pol, Volman, and Beishuizen]{van2010scaffolding}
Janneke Van~de Pol, Monique Volman, and Jos Beishuizen.
\newblock Scaffolding in teacher--student interaction: A decade of research.
\newblock \emph{Educational Psychology Review}, 22\penalty0 (3):\penalty0 271--296, 2010.

\bibitem[Williams et~al.(2024)Williams, Carroll, Narang, Weisser, Murphy, and Dragan]{williams2024targeted}
Marcus Williams, Micah Carroll, Adhyyan Narang, Constantin Weisser, Brendan Murphy, and Anca Dragan.
\newblock On targeted manipulation and deception when optimizing llms for user feedback.
\newblock \emph{arXiv preprint arXiv:2411.02306}, 2024.

\bibitem[Yeung \& Summerfield(2012)Yeung and Summerfield]{yeung2012metacognition}
Nick Yeung and Christopher Summerfield.
\newblock Metacognition in human decision-making: confidence and error monitoring.
\newblock \emph{Philosophical Transactions of the Royal Society B: Biological Sciences}, 367\penalty0 (1594):\penalty0 1310--1321, 2012.

\bibitem[Zhao \& Hawkins(2025)Zhao and Hawkins]{zhao2025comparing}
Haoran Zhao and Robert~D Hawkins.
\newblock Comparing human and llm politeness strategies in free production.
\newblock In \emph{Proceedings of the 2025 Conference on Empirical Methods in Natural Language Processing}, pp.\  16199--16227, 2025.

\bibitem[Zhi-Xuan et~al.(2025)Zhi-Xuan, Carroll, Franklin, and Ashton]{zhi2025beyond}
Tan Zhi-Xuan, Micah Carroll, Matija Franklin, and Hal Ashton.
\newblock Beyond preferences in {AI} alignment.
\newblock \emph{Philosophical Studies}, 182\penalty0 (7):\penalty0 1813--1863, 2025.

\bibitem[Zhou et~al.(2025)Zhou, Bao, Huang, Guo, Liang, Chen, Gao, Geyer, Moniz, Chawla, et~al.]{zhou2025emergent}
Yujun Zhou, Han Bao, Yue Huang, Kehan Guo, Zhenwen Liang, Pin-Yu Chen, Tian Gao, Werner Geyer, Nuno Moniz, Nitesh~V Chawla, et~al.
\newblock Emergent deceptive behaviors in reward-optimizing {LLMs}.
\newblock In \emph{Socially Responsible and Trustworthy Foundation Models at NeurIPS 2025}, 2025.

\end{thebibliography}
\end{document}